\documentclass[journal, twocolumn]{IEEEtran}
\usepackage{setspace}
\usepackage{amsmath,amsfonts}
\usepackage{algorithmic}
\usepackage{algorithm}
\usepackage{array}
\usepackage[caption=false,font=footnotesize,labelfont=rm,textfont=rm]{subfig}
\usepackage{textcomp}
\usepackage{stfloats}
\usepackage{url}
\usepackage{verbatim}
\usepackage{graphicx}
\usepackage{cite}
\graphicspath{{figure_set/}} 
\usepackage{subfig} 
\usepackage{float}
\usepackage{booktabs, tabularx, makecell, multirow}
\usepackage{bm}
\usepackage{amsmath}
\usepackage{amsfonts}
\usepackage{bbm}
\usepackage{amssymb}
\usepackage{ulem}
\usepackage{color}
\usepackage{dsfont}

\newcommand\Rt[1]{\textcolor{black}{#1}}

\usepackage{hyperref}
\hypersetup{hypertex=true,
	colorlinks=true,
	linkcolor=red,
	anchorcolor=blue,
	citecolor=green}

\begin{document}

\title{New Insights on Relieving Task-Recency Bias for Online Class Incremental Learning}

\author{Guoqiang Liang, Zhaojie Chen, Zhaoqiang Chen, Shiyu Ji, Yanning Zhang,~\IEEEmembership{Senior Member, IEEE}
\thanks{Manuscript received 15 February 2023; revised 10 May 2023; revised June 2023; revised 14 September 2023; accepted 14 October 2023. Date of publication XX XX 2023; date of current version XX XX 2023. This work was supported in part by the National Natural Science Foundation of China (No. U19B2037, No. 62376218, No. 61902321), the Ministry of Science and Technology Foundation funded project under Grant 2020AAA0106900, Natural Science Basic Research Program of Shaanxi (No. 2022JC-DW-08), and the Fundamental Research Funds for Central Universities of China under Grant G2021KY05104.}
\thanks{All authors are with Northwestern Polytechnical University, 710072, China. Guoqiang Liang, Zhaojie Chen, Shiyu Ji, Yanning Zhang are with National Engineering Laboratory for Integrated Aero-Space-Ground-Ocean Big Data Application Technology. Zhaoqiang Chen is with Key Laboratory of Big Data Storage and Management. Guoqiang Liang, Yanning Zhang, Zhaojie Chen and Zhaoqiang Chen are with School of Computer Science. Shiyu Ji is with School of Software Engineering. 
(e-mail: \{gqliang, ynzhang\}@nwpu.edu.cn; \{czjghost, jishiyu\_cs\}@163.com; chenzhaoqiang@mail.nwpu.edu.cn)}
}

\markboth{Journal of \LaTeX\ Class Files,~Vol.~14, No.~8, August~2021}%
{Shell \MakeLowercase{\textit{et al.}}: A Sample Article Using IEEEtran.cls for IEEE Journals}


\maketitle
\begin{abstract}
To imitate the ability of keeping learning of human, continual learning which can learn from a never-ending data stream has attracted more interests recently. In all settings, the online class incremental learning (OCIL), where incoming samples from data stream can be used only once, is more challenging and can be encountered more frequently in real world. Actually, all continual learning models face a stability-plasticity dilemma, where the stability means the ability to preserve old knowledge while the plasticity denotes the ability to incorporate new knowledge. Although replay-based methods have shown exceptional promise, most of them concentrate on the strategy for updating and retrieving memory to keep stability at the expense of plasticity. To strike a preferable trade-off between stability and plasticity, we propose an Adaptive Focus Shifting algorithm (AFS), which dynamically adjusts focus to ambiguous samples and non-target logits in model learning. Through a deep analysis of the task-recency bias caused by class imbalance, we propose a revised focal loss to mainly keep stability. \Rt{By utilizing a new weight function, the revised focal loss will pay more attention to current ambiguous samples, which are the potentially valuable samples to make model progress quickly.} To promote plasticity, we introduce a virtual knowledge distillation. By designing a virtual teacher, it assigns more attention to non-target classes, which can surmount overconfidence and encourage model to focus on inter-class information. Extensive experiments on three popular datasets for OCIL have shown the effectiveness of AFS. The code will be available at \url{https://github.com/czjghost/AFS}.

\end{abstract}

\begin{IEEEkeywords}
Online class incremental learning, Revised focal loss, Virtual knowledge distillation.
\end{IEEEkeywords}

\section{Introduction}
\Rt{Benefit from recent revolution of deep learning, image-based intelligent systems have achieved huge success in recent years. However, most of these impressive results are achieved by static models, which are trained using independent identically distributed datasets. If they are continually updated on non-iid data stream, catastrophic forgetting (CF) of old concepts will occur in high probability.} In contrast, human can learn concepts sequentially, where a complete loss of previous knowledge is rarely attested. To circumvent this challenge, Continual Learning, which can keep learning and accumulating knowledge from a never-ending data stream, has attracted more interests recently \cite{mai_online_survey,tcsvt_multi_domain,bhat2023task,22tpami_cil_survey}. 

Among different kinds of continual learning \cite{tcsvt_incre_meta,2023_cil_survey,tcsvt_domain_incre}, class incremental learning (CIL) is more practical \cite{three_scenerios_cl}. \Rt{Like previous methods \cite{gdumb,aaer,dvc,er_ace}, this paper further focuses on OCIL for image classification, where images from the data stream can be used only once. Compared with offline class incremental learning where samples from data stream can be used repeatedly, the online setting needs less memory and computation resource. Moreover, due to privacy concern, resource restrictions and unclear task boundary, it may not be feasible to store the whole dataset for a task.} Although there exist some methods based on regularization or parameter isolation \cite{agem,merlin}, the reply-based methods have dominated this area due to their better performance \cite{mai_online_survey}. Specifically, to alleviate catastrophic forgetting, a memory buffer is utilized to store a small number of samples from previous tasks, which are then combined with samples from current data stream to update models. Therefore, the main focus is how to update and retrieve samples from memory. For example, \cite{curiosity_cil} emphasized selective learning by quantifying the model's curiosity in each sample of new classes, but this method still needs a warm-up stage and works in offline class incremental learning. To relieve forgetting from the perspective of feature drift, \cite{aaer} proposed a novel AAER method, which contains drift-prevention and drift-compensation modules. In addition to forgetting, intransigence, the ability to incorporate new knowledge, is another key metric in continual learning \cite{ewc++}. To strike a better balance between stability and plasticity, the CIL methods should take them into account at the same time.

Although current methods obtain excellent performance, how to make full use of samples is usually neglected. Most of the models employ a cross-entropy loss, sometimes augmented with knowledge distillation (KD). However, in OCIL, there exists a severe class imbalance problem. In other words, samples for a new class are much more than that for an old class since the buffer can only store a small number of samples for each old class. Thus, if the model concerns each sample equally, the gradients of loss w.r.t to the weights in the final FC layer will be dominated by samples from new classes, which will lead to biased weights (i.e., task-recency bias). Moreover, due to the existence of lots of easy samples from new classes, the loss may tend to a local optimum by assigning a dramatically bigger value to the target logit. This can be achieved by learning some non-critical features. However, this feature may be not the key information to rightly classify an image, which will lead to CF when new tasks arrive. To alleviate this trend, focusing on valuable samples is an important way to achieve efficient updates as stated in many works \cite{mir,curiosity_cil,dvc,online_coreset}. \Rt{However, in OCIL setting, exactly measuring how valuable a sample for a model is tough. Meanwhile, these works always need much additional computational resource or the data distribution of current task.}


\Rt{Through a deep analysis of task-recency bias caused by the sample imbalance between new and old classes, we assume that each sample should be treated discriminatively. Specifically, we suggest paying more attention to potentially valuable samples, which the model can grasp better within limited time.} Therefore, we propose an Adaptive Focus Shifting algorithm (AFS), which can dynamically adjust the focus to current ambiguous samples. Specifically, the predicted score at the target class reflects the level of difficulty for a model to rightly classify a sample. Therefore, we split the model's predicted score into three intervals---easy sample interval (ESI), ambiguous sample interval (ASI) and hard sample interval (HSI), which can help us to find the potentially valuable samples from the perspective of current learning result. Then, based on curriculum learning \cite{curriculum}, we design a revised focal loss which always pays more attention to samples located at current ASI. By focusing more on these samples, the model will understand the current classes much better.  Furthermore, to strike a preferable trade-off between stability and plasticity, we propose virtual knowledge distillation. \Rt{By employing a virtual teacher, it encourages the model to attach importance to non-target classes, which can make model care more about inter-class information while surmounting overconfidence.}


To validate our framework, we have conducted extensive experiments on three popular datasets for OCIL, e.g. CIFAR-10 \cite{CIFAR}, CIFAR-100 \cite{CIFAR} and Mini-ImageNet \cite{MiniImageNet}. The experimental results have demonstrated that AFS strikes a preferable balance between stability and plasticity while achieving the best performance compared with the state-of-the-art methods.

Our main contribution can be summarized as follows:
\begin{itemize}	
	\item Through a deep analysis of the class imbalance problem, we emphasize that task-recency bias is the main challenge which arises from training each sample equally. Based on this, we propose a revised focal loss by concentrating more on current ASI samples to explicitly relieve task-recency bias.
	\item To avoid improving stability at the expense of plasticity while encouraging model to focus on inter-class information, we introduce a virtual teacher and propose a virtual knowledge distillation by utilizing its regularization role to pay more attention to non-target classes.	
	\item We have done extensive experiments on three popular benchmarks, whose results validate the effectiveness of ASF.	
\end{itemize}


\section{Related Work}
This section will review some closely related papers on continual learning, class imbalance and knowledge distillation.
\subsection{Continual Learning}
\subsubsection{Regularization Based Methods}
When training on each new task, regularization-based methods mainly regularize the model's parameters in order to protect some parameters that are important for previous tasks \cite{si,ewc,mas,ewc++}. EWC \cite{ewc} employs the diagonal of the Fisher Information Matrix to measure the importance of model parameters. Chaudhry et al. \cite{ewc++} propose the RWalk, which is faster and more efficient than EWC. Some methods \cite{gem,agem} focus on modifying the gradient directions that optimize the performance of new task while do not harm to old tasks. Though GEM \cite{gem} needs a memory buff, it mainly utilizes the samples of old classes to restrict the updating direction of gradients for new classes. To further complement previous works, SDC \cite{sdc} introduces an additional memory to store the prototypes of each class, which are used to compensate the feature drift of previous prototypes after training on a new task. Though regularization methods are elegant, they are not suitable for CIL scenario as discussed in \cite{lack_of_regularize}.

\subsubsection{Parameter Isolation Methods}
These methods allocate independent parameters to each task by extending the model's architecture or masking some irrelevant parameters. They could be parted into dynamic architecture \cite{lden,cndpm,treecnn} and fixed architecture \cite{packnet,merlin,itaml}. Though some recent works have achieved task-free, most of previous methods still need task indexes at inference time, which are unavailable in popular class incremental learning.

\subsubsection{Replay-based Methods}
Replay-based methods utilize a memory buffer to store a small number of samples from previous tasks. When a sample batch of current task comes, another batch will be retrieved from the memory buffer. Then these two batches are combined to update models. Compared with other two kinds, these methods are more efficient for OCIL scenario \cite{aser}. 

Current researches for replay-based methods focus on the strategy of memory updating, memory retrieving and model optimization. For instance, MIR \cite{mir} aims to retrieve the samples whose prediction will be the most negatively impacted by the foreseen parameters update. It has been proven that MIR is a simple but effective method under OCIL setting \cite{mai_online_survey}. GSS \cite{gss} concentrates on the strategy for memory updating and tries to make the gradient directions of samples diversified in memory buffer. However, it is computationally costly. ER \cite{er} apples the reservoir algorithm to update memory buffer and randomly retrieves samples from the memory buffer. Due to its excellent performance, it has become a backbone for many replay-based methods. ASER \cite{aser} focuses on both strategies and adopts an adversarial shapley value scoring method to estimate the ability of a sample to maintain learning stability and encourage plasticity. Bhat et al. \cite{bhat2022consistency} propose to cast consistency regularization as a self-supervised pretext task. Although GDumb \cite{gdumb} is not elaborate for continual learning, it shows strongly competitive performance. Concretely, it trains a model from scratch using all samples in the memory buffer before each inference and designs a class-balanced sampler to replace the reservoir algorithm. Recently, DVC, OCM and SCR \cite{dvc,ocm,scr} commit themselves to design a suitable rehearsal loss, which instructs the model to better learn from the non-i.i.d data stream.

Our method belongs to the replay-based category. However, different from the above methods, we mainly focus on the optimization objective to alleviate task-recency bias when using the combination of FC layer and cross-entropy loss.

\begin{figure*}[ht]
	\centering 
	\subfloat[The mean weights in the FC layer for old and new classes.]{\includegraphics[scale=0.24]{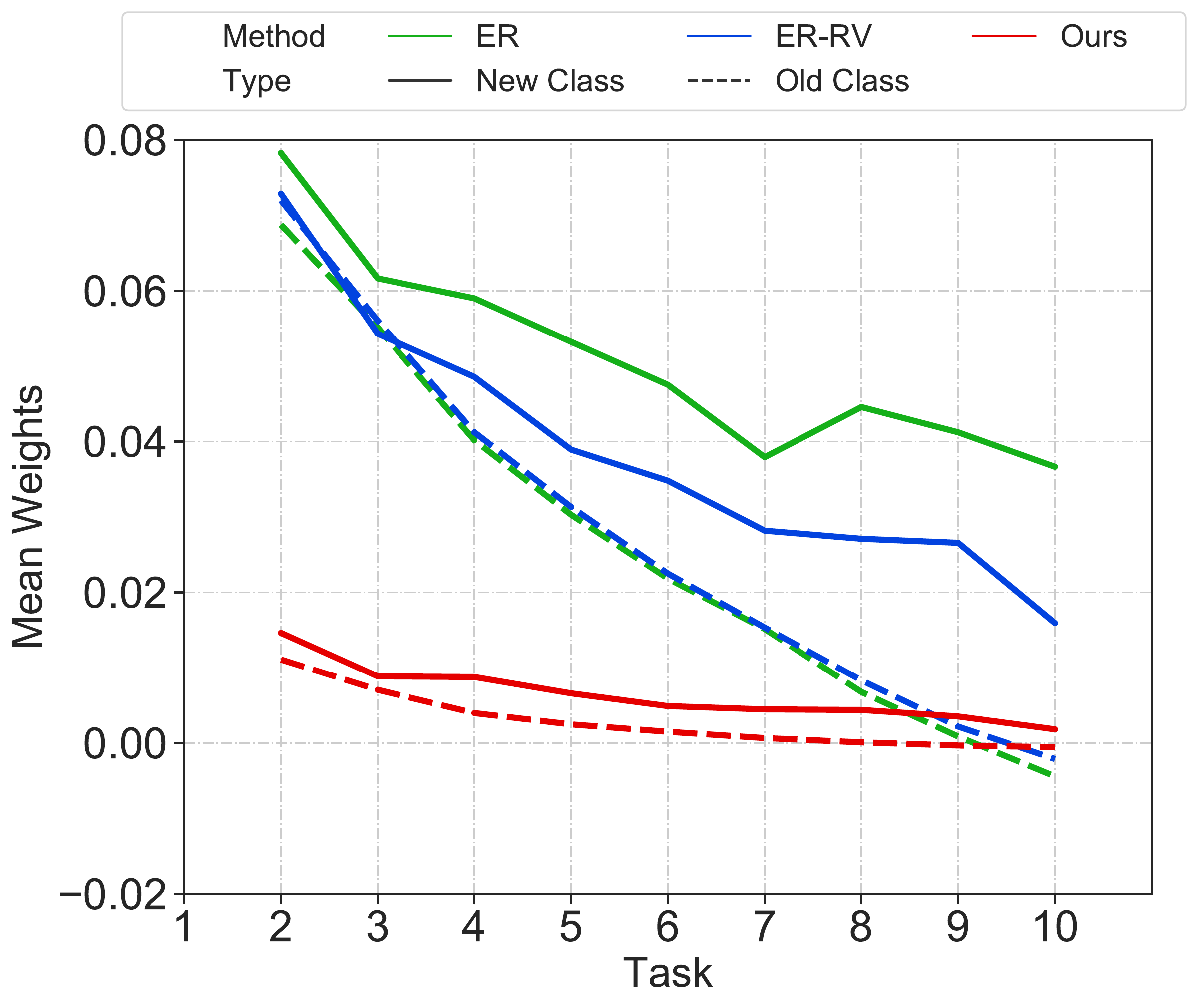}
		\label{mean_weight}}
	\subfloat[The mean logits for old and new classes.]{\includegraphics[scale=0.24]{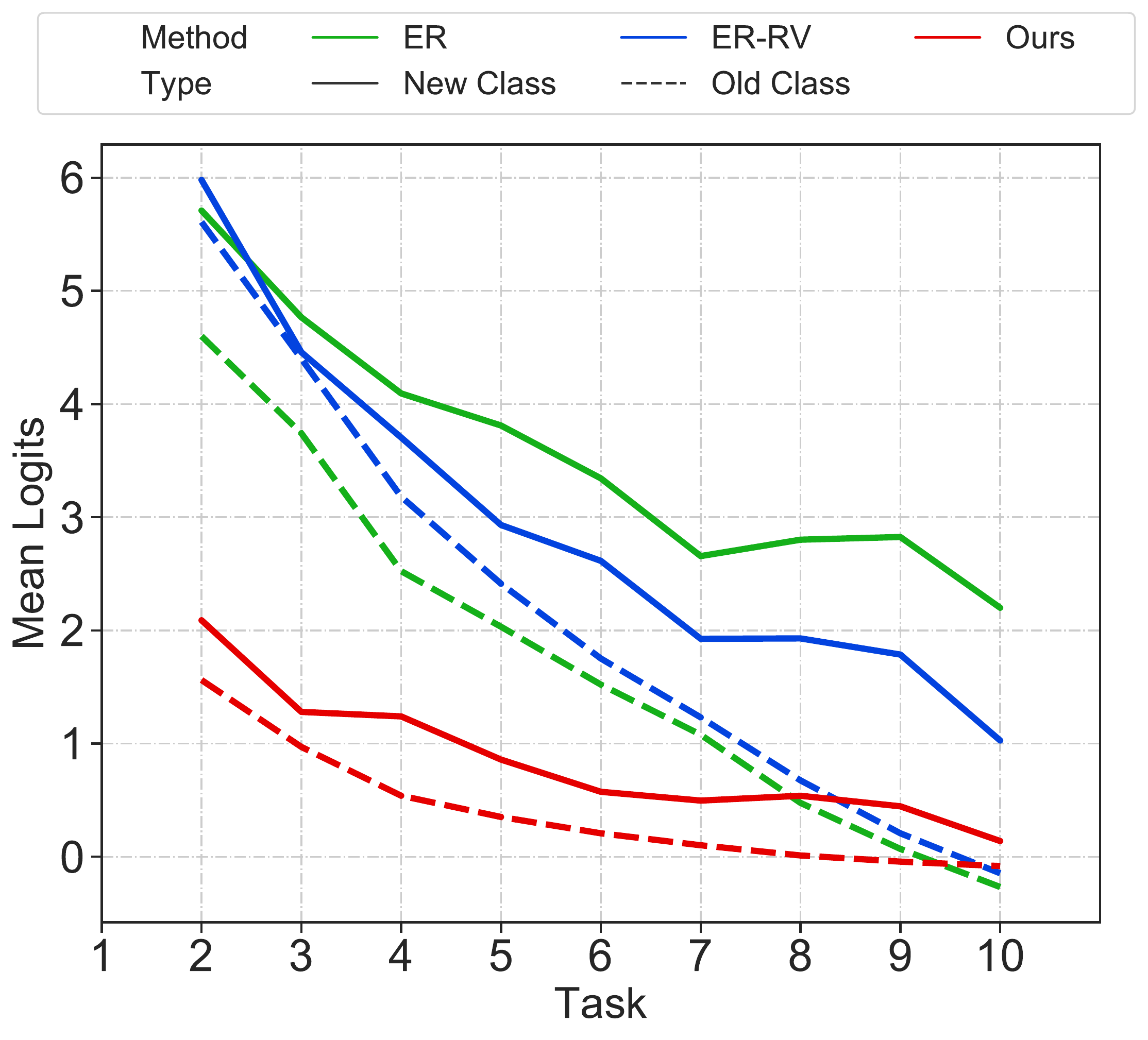} 
		\label{mean_logit}}
	\subfloat[The number of samples with different difficulty for new classes in the training set after each task.]{\includegraphics[scale=0.24]{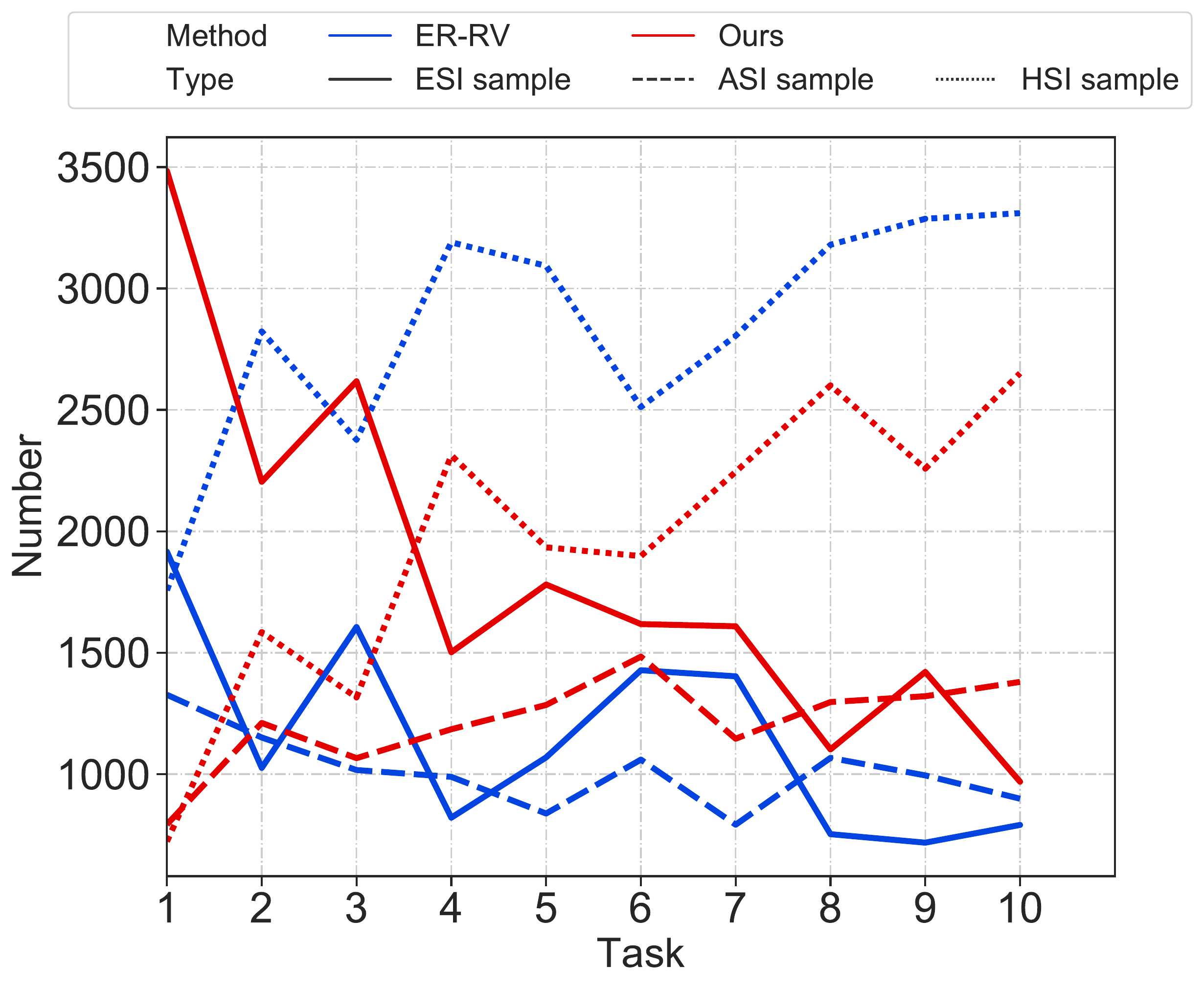} 
		\label{sample_type}}
	\caption{The comparison of task-recency bias for ER, ER-RV and AFS on CIFAR-100 with 5k memory. Note that the mean weights in the FC layer contain the bias item while the mean logits of new class and old class include both target logit and non-target logits when calculating the mean value.}
	\label{mean_weight_and_logit}
\end{figure*}

\subsection{Class Imbalance} 
Due to the limited samples for old classes, class imbalance is a serious problem in CIL, which is one of the most important factors for CF \cite{bic}. To alleviate this problem, multiple strategies have been proposed. For example, \cite{ss_il} employs two different Softmax output layers for new and old classes respectively. \cite{eeil} first designs an extra fine-tuning step using a balanced subset from memory and current task. \cite{ber} further develops a simplified version called review trick (RV), which is also used in SCR \cite{scr}. Replacing the combination of FC and soft-max classifier with the nearest class mean classifier (NCM) is another popular option \cite{sdc,scr}. Besides, the class imbalance occurs in other tasks, such as long-tailed classification and object detection \cite{oltr,eql,efl}. In this paper, a revised focal loss is developed to tackle this problem, which is further combined with the widely used RV. 

\subsection{Knowledge Distillation}
Knowledge distillation \cite{kd} has shown impressive achievement for knowledge transfer. LwF \cite{lwf} first introduces KD into continual learning, which gets better performance. From then on, different ways have been proposed to combine KD with classification loss for continual learning \cite{eeil,lucir}. For example, LUCIR \cite{lucir} applies the KD loss to the feature level, which provides a strong constraint on the previous knowledge. However, Zhao et al. \cite{wa} point out that the positive effect of KD was hidden and limited since the task-recency bias was increasingly accumulated when training tasks one by one. Moreover, though KD loss has been widely used in continual learning, only a few works \cite{il_online_scinerio} applied it to OCIL. Besides, it always needs clear task boundary during training. Here, we introduce a virtual KD loss to OCIL, which plays a regularization role with no need for clear task boundary and task identity during training.


\section{Methodology}
In this section, we first introduce the problem definition of OCIL. Then, based on a deep analysis of the task-recency bias caused by class imbalance problem, we will detail the proposed method. 

\subsection{Problem Definition}
Following the recent literatures \cite{mir,aser,mai_online_survey}, we consider the online supervised class incremental learning, where a model needs to learn new classes continually from an online non-i.i.d data stream. For convenience, we represent the data stream as $D = \{ D_{1}, D_{2}, ..., D_{N} \}$ over $X \times Y$, where $X$ are input samples, $Y$ are their labels, and $N$ denotes the number of total tasks. Specially, the class set of each task is unique (i.e. $Y_i \cap Y_j=\emptyset$ for $\forall \ i, j \in \{1,...,N \}, i \neq j$). The dataeset of a task $\mathcal{T}$ contains a series of sample batches, $D_{\mathcal{T}} = \{ B_{1}, B_{2}, ..., B_{b_{\mathcal{T}}} \}$, where $b_{\mathcal{T}}$ is the total number of batches. And $B_{k} = \left\{ \left(x_i, y_i \right) \right\}_{i=1}^{n_k}$, where $x_i$ is an input sample, $y_i$ is its corresponding class label. At task $\mathcal{T}$, the old classes refer to the classes $y \in Y_j, j<\mathcal{T}$ (denoted as $C_{old}$) while the new classes mean $y_i \in Y_{\mathcal{T}}$ (denoted as $C_{new}$). Besides, the model consists of a feature extractor $f$ parameterized by $\theta$ and a classifier $g$. Like \cite{ewc++,mir,er_ace}, we adopt the single-head evaluation setup, where the task identity is not available during inference. Note that the data stream can be seen only once, which means the data $D_{\mathcal{T}}$ can be used to train the model for one epoch.

\subsection{Task-recency Bias}
In OCIL, due to the limited memory size, the samples from old classes are much more than that for new classes,  which is also called class imbalance. \Rt{Next, we retrospect how does it lead to biased weights in the last FC layer, which are the main cause for catastrophic forgetting according to recent works \cite{ss_il,scr,mai_online_survey,22tpami_cil_survey}.}

We use $\bm{W_k} \in \mathbb{R}^{d}$ to denote the weights corresponding to class $k$ in the final FC layer, i.e. the classifier $g$. For simplicity, the $\bm{W_k}$ has included the bias item. Thus, for an input sample $x_i$, the predicted logit for class $k$ is $z_k = \bm{W_k} \cdot f\left( x_i \right) $. Then, we can get the gradient of the cross-entropy loss w.r.t $\bm{W_k}$:
\begin{align}
	\frac{\partial CE}{\partial \bm{W_k}} &=\frac{\partial CE}{\partial z_k} \frac{\partial z_k}{\partial \bm{W_k}} \nonumber \\
	&=\left(  \frac{e^{z_k}}{ \sum_{j \in C_{old} + C_{new}} e^{z_j} } -  \mathds{1}_{\left\{ k=y_i \right\}} \right) f_{\theta} \left( x_i \right),
	\label{CE_W}
\end{align}
where $y_i$ is the corresponding ground-truth class, $\mathds{1}_{\left\{ \cdot \right\}}$ is an indicator function. Since the ResNet \cite{resnet} is generally used as the feature extractor, the second term $\frac{\partial z_k}{\partial \bm{W_k}} = f_{\theta} \left( x_i \right)$ is always positive because of the ReLU activation unit. So we mainly consider the first term $\frac{\partial CE}{\partial z_k}$. When $k \neq y_i$, the gradient will be positive. Thus, the weights will be continually penalized according to the rule of gradient descent $\bm{W_k} = \bm{W_k} - \lambda \cdot \frac{\partial CE}{\partial \bm{W_k}}$, where $\lambda > 0$ is the learning rate. 

Specially, for class $k \in C_{old}$, the situation $k \neq y_i$ will occur much more frequently since the number of samples for each old class is much smaller than that of a new class. In other words, the weights corresponding to old classes will receive more penalties than rewards compared to the weights for new classes, which means the weights of old classes are kept to decrease. In contrast, the weights for new class will keep increasing because of more rewards. Thus, the weights of new class overwhelm that for old classes. Consequently, the logits of old classes tend to become smaller and the model is prone to classify input samples to new classes. Moreover, compared with old classes, the new class will contribute to many easy samples generally under training over abundant samples. When using the cross-entropy loss and one-hot label, the final goal is to make the target logit output remarkably distinctive, this will cause excessive updating in the weights of new class and aggravate task-recency bias.

Furthermore, our initial experiments also validate this bias when using the combination of FC layer and soft-max classifier. In Fig. \ref{mean_weight_and_logit}\subref{mean_weight} and \subref{mean_logit}, we show the mean weights of FC layer and the predicted logits for some typical methods ER and ER-RV. As shown in this figure, the mean weights in the FC layer for new classes are much larger than that of old classes. And the model predicts a larger logit for new classes than old classes. Moreover, the margin becomes bigger when there are more tasks. These results reveal that there exists a severe task-recency bias in the popular replay-based framework ER.

\subsection{Revised Focal Loss}
As stated above, the class imbalance problem will lead to severe task-recency bias, which further causes CF. This class imbalance problem originates from the very small amount of samples for each old class. Due to the restriction of memory, it is infeasible to store more samples. One of the solutions is to store samples with higher value. To achieve this goal, many memory management strategies  have been proposed \cite{mir,curiosity_cil,dvc,online_coreset}, which are to select/store several most valuable samples. Through learning these samples, the model can learn better features to mitigate CF. \Rt{However, it is very tough to exactly measure the value of a sample, which often needs many additional computational resources or the whole distribution of current task.} Besides, this class imbalance problem is a little similar to the imbalance of samples for rare and frequent classes in object detection, which can be alleviated by focal loss \cite{focal}. Specifically, it increases the weights of hard samples which are almost from rare classes while decreasing weights for other samples which are from frequent classes.

On the other hand, to improve learning efficiently, humans in different stages need to do exercises with various difficulty and sometimes further focus on fallible questions. To imitate this process, curriculum learning \cite{curriculum}, a strategy to efficiently train a model, advocates to let the model first learn easy samples, and gradually advance to complex samples. Inspired by these, we propose the model should learn new classes gradually and focus on fallible samples. Specifically, in our opinion, various samples offer diverse value at different learning stages when a model masters new knowledge gradually. Since the model can not see a class multiple times, we should dynamically adjust its focus and make it pay more attention to important samples according to its current ability. 

Before introducing our method, we first introduce how to measure the difficulty of samples. Specifically, we split the predicted target logit after Softmax (i.e. the predicted score for ground-truth class) into three intervals, which loosely measure the value of samples for the current model. For an input sample $x_i$, we denote the predicted target score as $p_t = e^{z_t} / {\sum_{j=1}^{C} e^{z_j}}$, where $t = y_i$ is the ground-truth class. Empirically, we think $p_t \in \left[ 0.0 , 0.3  \right)$ as the hard sample interval (HSI), $p_t \in \left[ 0.3 , 0.6  \right]$ as the ambiguous sample interval (ASI) and $p_t \in \left( 0.6, 1.0 \right]$ as the easy sample interval (ESI). Note that the boundary of each interval is relatively loose and these intervals just introduce a reference for classification of samples to better understand the focus of our weight function.

\subsubsection{\textbf{Focus on What to Learn}}
Which type of samples are more valuable? In focal loss, hard samples are considered to be the most valuable \cite{focal}. Compared with this traditional object recognition task, the model in OCIL needs to continually learn new classes while most samples for old classes are not available. Actually, hard samples in OCIL arise from two aspects. One is the fake hard samples, which always are the first dozen of batches from new classes. Since the model never sees these classes, they will be given lower confidence. Once the model has understood these classes after several batches, these hard samples will disappear. The other is the real hard samples. \Rt{Considering the small number of old samples, most of hard samples belong to new classes (both real and fake ones).} As stated in the above subsection, the task-recency bias will be aggregated if the model focuses more on these hard samples of new classes all the time. On the other hand, even if they are grasped, a much tighter classification boundary will be generated. To maintain this strict boundary, more samples are usually needed since deep learning models depend on the data distribution heavily. \Rt{However, this is impossible in OCIL and the tighter boundary will be changed in a very high probability when a new class arrives.} Thus, focusing on hard samples is not suitable for OCIL.

For ambiguous samples, the model may understand them partially through some local features. In detail, due to the existence of similar classes, the model can not distinguish them confidently and thus assign relatively lower score for target class, meaning that there exist important features between target class and non-target classes, which the model grasps poorly. Furthermore, many theories of education suggest that the learning process is optimal when an appropriate level of challenge is maintained \cite{education_psychology}. If a task is too easy, there is little to learn. If a task is too difficult, the learner can be overwhelmed \cite{challenge_point}. As a result, we emphasize the model should focus on current ambiguous samples \Rt{(i.e. potentially valuable samples in our method)} according to its prediction.

\subsubsection{Revised Focal Loss}
Based on the above analysis, we propose a Revised Focal Loss (RFL), which shifts the focus to ASI samples.  Specifically, it could be defined as 
\begin{equation}
	RFL \left( p_{t} \right) = -\alpha e ^ { - \frac {\left(  p_{t} - \mu \right)^{2}}{\sigma}  }  \log \left( p_{t} \right),
	\label{rfl_loss}
\end{equation}
where $\alpha$ is a hyper-parameter as in the focal loss, $\mu$ and $\sigma$ are two new hyper-parameters. $\mu$ determines which kinds of difficult samples should be focused on while $\sigma$ controls the extent of decline on both sides of $\mu$. As stated previously, we need to concentrate on ambiguous samples. Therefore, we empirically set $\mu$ to $0.3$, which can satisfy the requirement. Through Eq. (\ref{rfl_loss}), we can re-balance the gradients between ASI samples and other types. 

The original focal loss could be denoted as 
\begin{equation}
	FL \left( p_{t} \right) = -\alpha \left( 1 - p_{t} \right)^{\gamma} \log \left( p_{t} \right).
\end{equation}
It down-weights the loss of easy samples and focuses on the learning of hard samples through the focusing parameter $\gamma$. Compared with this, we design a new weight function, whose focus is ambiguous samples. 



\subsubsection{The difference between FL and RFL} We further compare the cross entropy loss (CE), FL and RFL from the aspect of loss value, weighting scheme and gradient for target logit. For convenience, we first give the formulation of widely used CE:
\begin{equation}
	CE \left( p_{t} \right) = -\log \left( p_{t} \right).
\end{equation}

\begin{figure}[t]
	\centering
	\subfloat[Loss curve over various difficulty sampels.]{
		\includegraphics[width=0.6\linewidth]{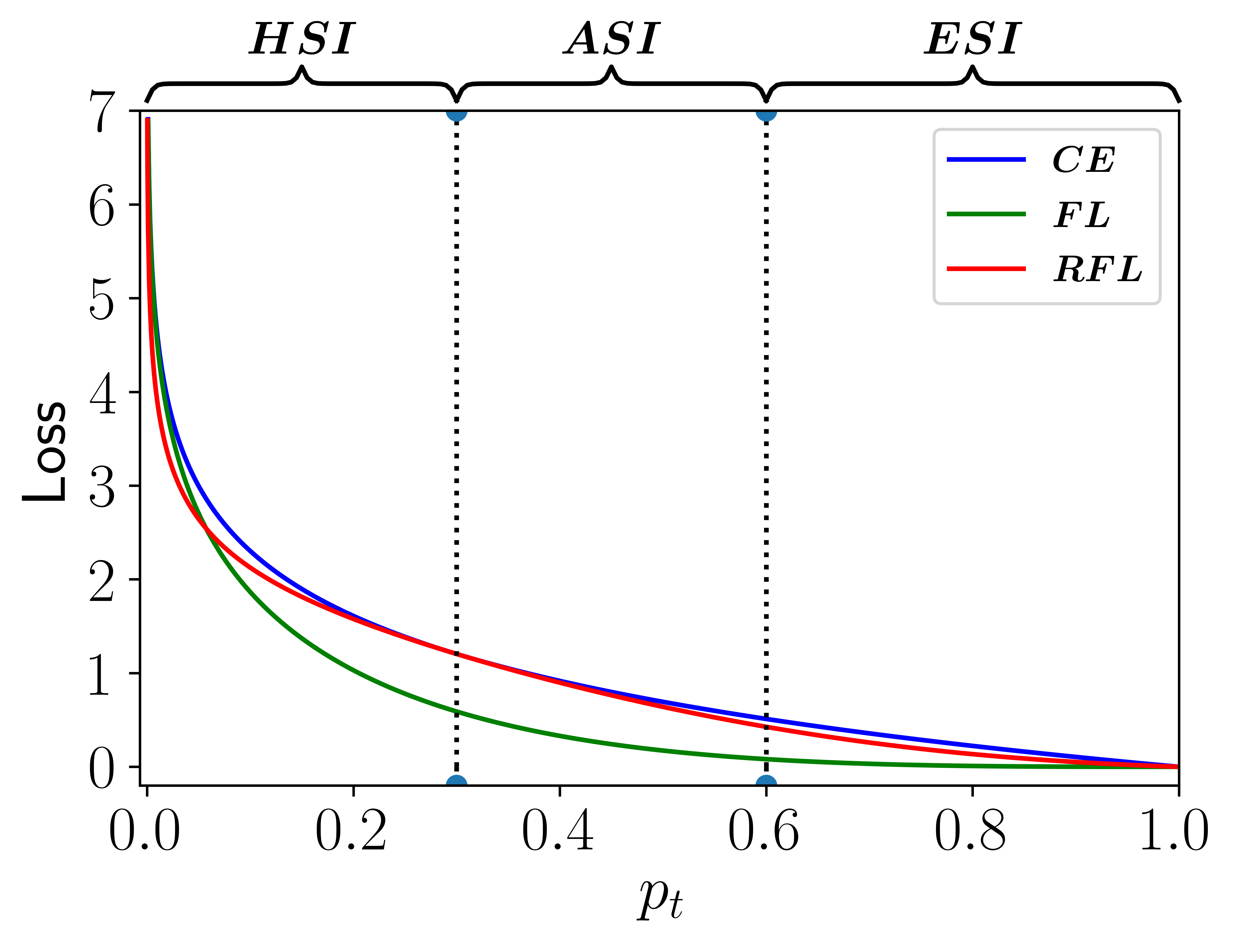}
		\label{loss}
	}
	\quad
	\subfloat[Weighting schemes.] {
		\includegraphics[width=0.48\linewidth]{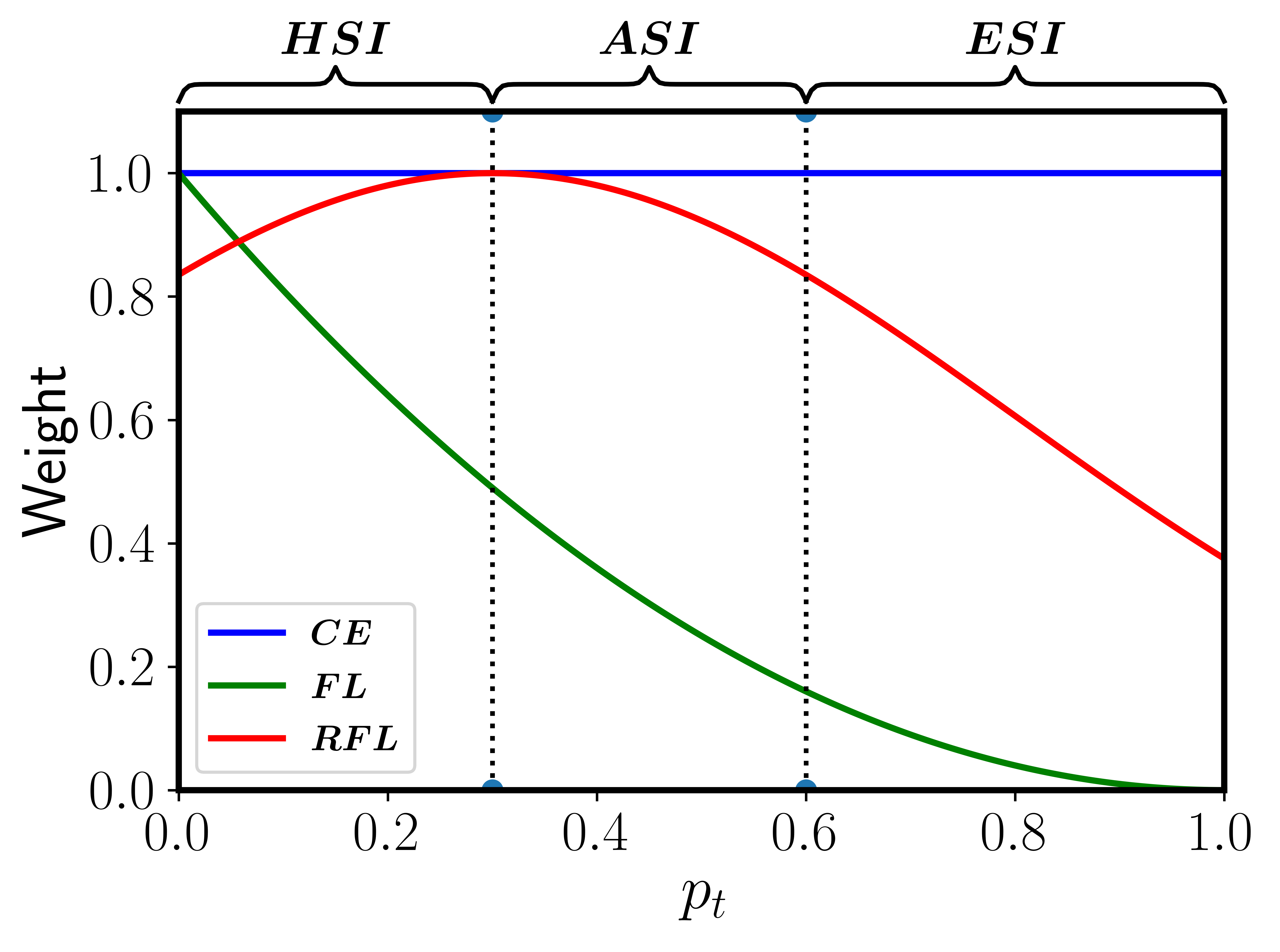}
		\label{weight}
	}
	\subfloat[Gradients w.r.t target logit.] {
		\includegraphics[width=0.50\linewidth]{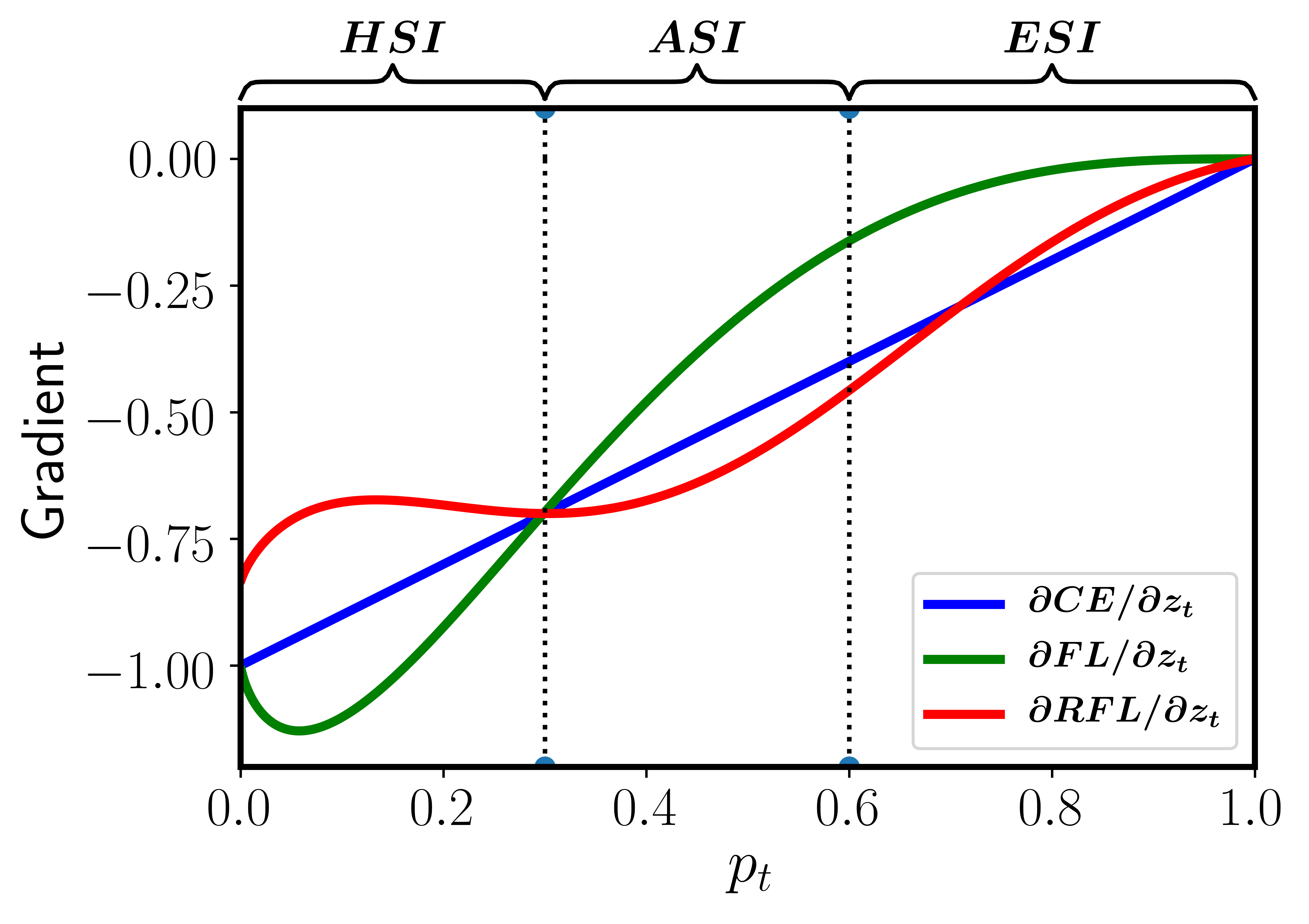}
		\label{target_derivates}
	}
	
	\caption{Comparison of loss curve, weighting schemes and gradients of different loss functions. \Rt{Note that $\gamma = 2$, $\mu=0.3$, $\sigma=0.5$. We set $\alpha=1.0$ for convenient comparison}.}
	\label{loss_and_weight}
\end{figure}

With all the definitions, we then analyze the variation of loss and weighting scheme w.r.t the difficulty of samples in three loss functions, where $p_t$ is employed to measure the difficulty of samples. \Rt{In Fig. \ref{loss_and_weight}\subref{loss}, we can see that the FL loss for easy and ambiguous samples declines more quickly compared with CE and RFL, which means FL focuses on hard samples. Compared with CE, our RFL only declines the loss for both ESI and HSI samples. Moreover, when the predicted target score increases, the weight of samples in RFL first increases and then reduces after the peak point as shown in Fig. \ref{loss_and_weight}\subref{weight}. Although the weight for ASI and HSI are reducing, the speed are different, which produces different effect in Fig. \ref{loss_and_weight}\subref{target_derivates}. This weight function imitates the human learning process---Human tends to focus on learning the knowledge which is more likely to achieve big progress within limited time.}

	



We further analyze the gradient for deeper understanding. Note that $\frac{\partial z_k}{\partial \bm{W_k}}$ is always positive when using ReLU. For simplicity, we just compare the derivatives of different losses w.r.t the ground-truth logit $z_t$ (the non-target logit shows similar results). Specifically, they can be denoted as:
\begin{equation}
	\frac{\partial CE}{\partial z_t} = -\left( 1 - p_t \right) = - Q,
\end{equation}
\begin{equation}
	\frac{\partial FL}{\partial z_t} = Q^{\gamma} \left( \gamma p_t \log p_t + p_t - 1 \right),
\end{equation}
\begin{equation}
	\frac{\partial RFL}{\partial z_t} =  e ^ { - \frac{\left( p_t - \mu \right) ^ 2}{\sigma} } \left( \frac{2 p_t \left( p_t - \mu \right) \log p_t}{\sigma} - 1\right) Q.
\end{equation}

\Rt{The curves of these derivatives are shown in Fig. \ref{loss_and_weight}\subref{target_derivates}, where the $p_t$ at the first intersection is around 0.3. Compared with CE, FL promotes the absolute value of gradients (AVG) of HSI samples while our RFL reduces them. For samples whose probability are higher than 0.3, FL directly reduces their AVG. In contrast, our RFL increases the AVG of ASI samples and decreases that of ESI samples. By doing so, we reduce the influence of HSI and ESI samples while promoting the impact of ASI samples during model training. Hence, by paying more attention to ASI samples, we make the classification results polarized (i.e. samples belong to either HSI or ESI, instead of locating at ASI).}

\Rt{Different from previous methods \cite{gss,aser}, which are to select and store several most effective samples to preserve the distribution of old classes, our measurement of sample's value is to re-weight samples to solve the class imbalance problem. Moreover, our method does not need additional computation resources by directly employing the predicted probability obtained in the model's forward process.}

\subsection{Virtual Knowledge Distillation}
Since a general viewpoint of continual learning  is a stability-plasticity dilemma \cite{cl_survey}, the plasticity referring to the ability of integrating new knowledge is another key metric. For each class of a task, there are lots of samples, many of which are usually easy samples after several model training iterations. Therefore, the model may tend to be over-fitted to these easy samples. \Rt{In other words, the model will learn a tight boundary. However, in OCIL, the inter-class relationship is changing over time since new classes are arriving continually. The tight classification boundary for current classes is easily affected by new classes in the future tasks.} To leave space for integrating new knowledge, overconfidence should be avoided. Specifically, we should pay attention to non-target classes to learn more discriminative features. 

To improve plasticity, a naive way is label smoothing regularization (LSR) \cite{labelsmooth}, which lets the model escape from overconfidence. However, our initial experiments deny this since it can not control the attention rate on non-target classes. Therefore, we turn to knowledge distillation. Inspired by \cite{rkd_lsr}, we introduce a virtual teacher to promote plasticity by utilizing the regularized role of knowledge distillation. 


\subsubsection{The Limitation of KD for Online CIL} According to \cite{kd}, the common KD for a teacher-student model can be formulated as 
\begin{equation}
	L_{kd} = -\sum_{i=1}^{C} \hat{p_i}^T \log p_i^T,
\end{equation}
where $\hat{p_i}^T, {p_i}^T$ are the soften predicted probability of $i-th$ class by the teacher and student respectively, $C$ is the total number of classes. Generally, ${p_i}^T$ is converted from the predicted logit using soft-max with temperature $T$. 

In CIL, this loss has been used to distill the knowledge of old models trained using previous tasks \cite{eeil,bic,wa}. Although KD is very popular in CL, its positive effect is hidden and limited due to the class imbalance problem. The task-recency bias is increasingly accumulated when training tasks one by one \cite{wa}. Moreover, traditional methods use KD to alleviate forgetting instead of intransigence. However, some experimental results \cite{mai_online_survey} show that KD only seems to be conducive to methods whose performance are pretty poor. And \cite{il2m} pointed out that KD is in conflict with balanced fine-tune which the review trick develops from. The teacher model contains biased knowledge while the balanced fine-tune can learn actual knowledge among current all classes to alleviate CF. On the other hand, since the task boundary is not clear most time in real-world scenes, how to define old classes and choose the teacher model become a headache.



\begin{figure}[t]
	\centering
	\includegraphics[width=0.7\linewidth]{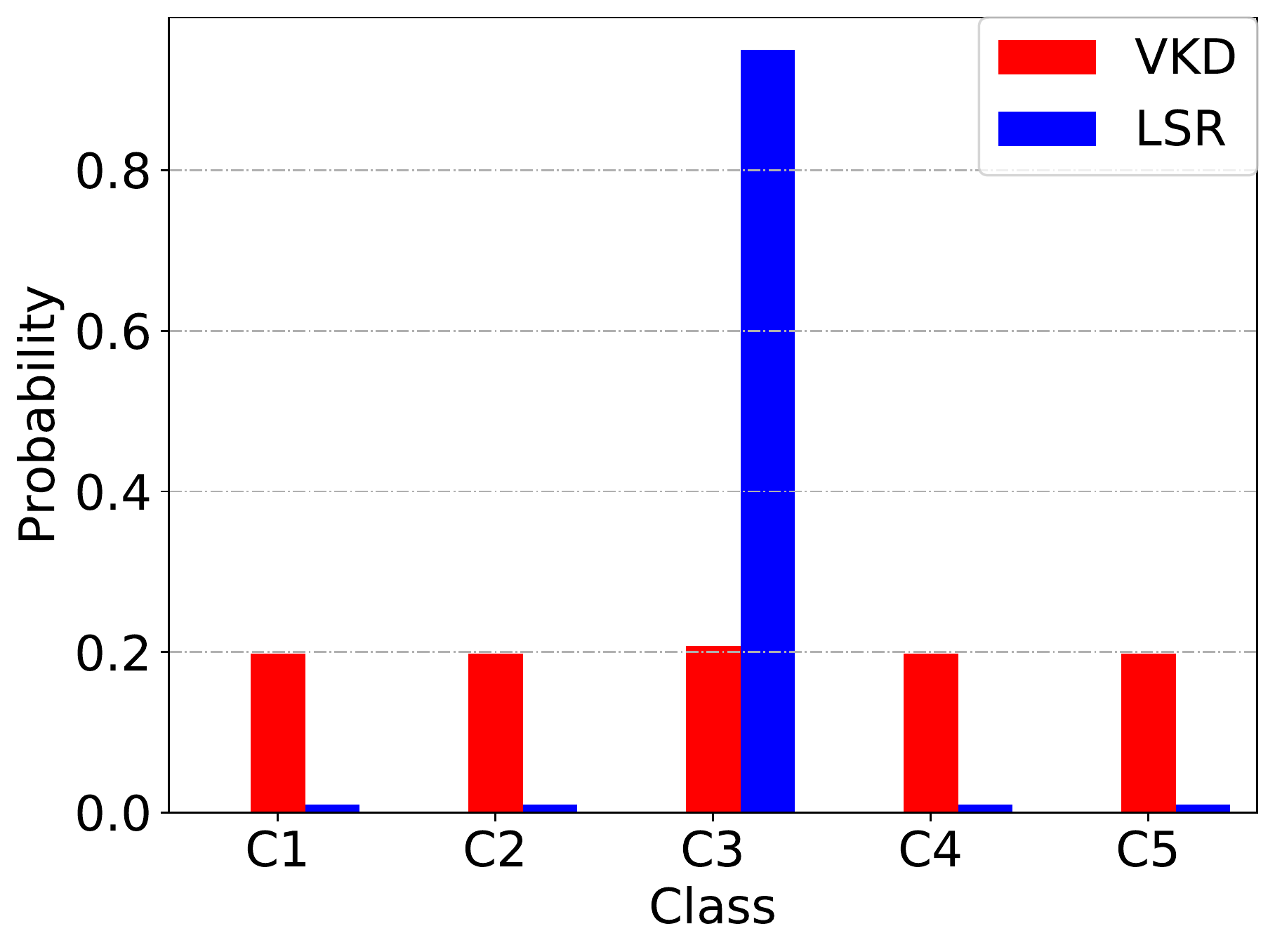}
	\caption{\Rt{The label distribution of LSR and VKD in temperature $T=20$. 'C3' is the ground-truth}.}
	\label{diff_lsr_vkd}
\end{figure}

\subsubsection{Knowledge Distillation via Label Smoothing Regularization}
To address the above problem, we introduce a virtual knowledge distillation (VKD) to OCIL. Instead of using  logits from the model updated on the previous task, we design a virtual teacher, which produces a virtual logit $v_i$ as following
\begin{equation}
	v_i =  \left\{\begin{array}{ll}1-\epsilon & \text{if } i = t, \\
		\epsilon / \left( C-1 \right)    &  \text{otherwise}, \\
	\end{array} \right.
\end{equation}
where $\epsilon$ is a small constant and $C$ indicates the total number of classes. According to the current research, $C$ is a known value in advance. We leave its dynamic update for future work. With this definition, the corresponding probability for class $i$ could be formulated as $q_i^T = e^{ \left( v_i / T \right)} / \sum_{j=1}^{C} e^{\left( v_j / T \right)}$. Here, we also use a high temperature scale $T$. As stated in \cite{rkd_lsr}, KD is interpreted as a regularization method. Their experimental results show that KD via label smoothing regularization can get competitive performance compared with traditional KD and outperform label smoothing regularization. Hence, our way for generating the virtual logits adopts the label smoothing. From the regularization perspective of VKD, the virtual teacher offers a random noise to encourage the model to output a softer distribution of probability (similar to LSR). \Rt{In contrast, the cross entropy loss encourages the output logits dramatically distinctive, which may potentially lead to over-fitting} \cite{bagtrick_lsr}. 


Since we only design a new way to generate logits, the derivative of the VKD loss w.r.t. a logit can be approximated like \cite{kd} 
\begin{equation}
	\frac{\partial L_{vkd}}{\partial z_i} \approx \frac{1}{N T^{2}} \left( z_i - {v_i} \right).
	\label{kd_derivative}
\end{equation}

\Rt{As shown in Eq. \eqref{kd_derivative}, our VKD still encourages the model to imitate the output of virtual teacher (like KD), which is smoothed ground-truth. }

\Rt{In the level of model optimization, both VKD and LSR are to alleviate overconfidence. Specifically, they aim to output the largest logit for target class while leave small value for non-target classes rather than zero value. However, there exists a significant difference. Since we usually use high temperature (for example $T=20$), the difference of target and non-target labels decreases largely in VKD. From the example in Fig. \ref{diff_lsr_vkd}, the labels for non-target classes are much larger than that in LSR, even close to the target label, which means stronger smoothing property. Hence, VKD encourages the model to concern inter-class relationships, which can enhance model's ability to learn discriminative features. Besides, the virtual teacher in VKD can be assumed as an ideal teacher with very high confidence. Hence, using high temperature will weaken this teacher so that student can learn gradually. Otherwise, it may force the student to learn an ideal output which may influence its plasticity. In a word, VKD contains both the strong smoothing property (high temperature) and more proper distribution (less overconfidence, same to LSR).} Hence, VKD can augment the ability to quickly incorporate new knowledge, which can improve plasticity without introducing task-recency bias. Different from previous KD which can be only used to old classes, our VKD does not need previous outputs. Therefore, we apply the VKD to all samples from all tasks.



In all, the final loss  $\mathcal{L}$ can be formulated as:
\begin{equation}
	\mathcal{L} = -  \alpha e ^ { - \frac {\left(  p_{t} - \mu \right)^{2}}{\sigma}  }  \log \left( p_{t} \right)- \beta T^2\sum_{i=1}^{C} q_i^T \log p_i^T,
	\label{final_loss}
\end{equation}
where $\beta$ is a hyper-parameter controlling the strength of regularization by VKD. Algorithm \ref{algorithm_ours} summarizes our training procedure while our inference process can be conducted as traditional cross-entropy methods.

\section{Experiment}
To validate the proposed method, this section will compare it with some newly published works on three popular datasets for OCIL. We first introduce the benchmark datasets, metrics, baselines and implementation details. Then, we will report and analyze the results. 

\begin{algorithm}[t!]
	\caption{AFS Training Algorithm.}
	\label{algorithm_ours}
	\begin{algorithmic}
		\STATE \textbf{Input}: Data Stream $D$, Memory $\mathcal{M}$, Learning Rate $\lambda$, Learning Rate for RV $\lambda_{RV}$, Data Augmentation $Aug(\cdot)$, Memory Size $M$
		\STATE \textbf{Output}: Optimized Network Parameters $\theta$
		\STATE \textbf{Initialize}: Memory $\mathcal{M} \leftarrow \left\{\right\}*M$, Initialize Network Parameters randomly $\theta$, Counter $tot \leftarrow 0$
	\end{algorithmic}
	\label{alg}
	\begin{algorithmic}[1]
		\STATE \textbf{for} $n \in {1,...,N}$ \textbf{do} {
			\STATE \hspace{0.5cm} // \textit{Learning New Task}
			\STATE \hspace{0.5cm} \textbf{for} $B_{k} \sim D_n$ \textbf{do}{
				\STATE \hspace{1.0cm} $B_{\mathcal{M}} \leftarrow RandomRetrieval\left( \mathcal{M} \right)$
				\STATE \hspace{1.0cm} $B_{\mathcal{A}} \leftarrow Concat([B_{\mathcal{M}}, Aug(B_{\mathcal{M}})])$
				\STATE \hspace{1.0cm} Compute $\mathcal{L}$ by equation \eqref{final_loss} for $B_{k} \cup B_{\mathcal{A}}$
				\STATE \hspace{1.0cm} $\theta \leftarrow SGD\left(\mathcal{L},\theta,\lambda\right))$ 
				\STATE \hspace{1.0cm} $\mathcal{M} \leftarrow ReservoirUpdate(B_{k},\mathcal{M},tot)$
				\STATE \hspace{1.0cm} $tot \leftarrow tot + |B_{k}|$
			}
			\STATE \hspace{0.5cm} \textbf{end for}
			\STATE \hspace{0.5cm} // \textit{Review Process}
			\STATE \hspace{0.5cm} \textbf{for} $B_{j}^{RV} \sim \mathcal{M}$ \textbf{do}
			\STATE \hspace{1.0cm} Compute $\mathcal{L}_{RFL}$ by equation \eqref{rfl_loss} for $B_{j}^{RV}$
			\STATE \hspace{1.0cm} $\theta \leftarrow SGD\left(\mathcal{L}_{RFL},\theta,\lambda_{RV} \right))$ 
			\STATE \hspace{0.5cm} \textbf{end for}
		}
		\STATE \textbf{end for}
		\STATE \textbf{return} $\theta$
	\end{algorithmic}
	\label{alg2}
\end{algorithm}

\subsection{Datasets and Metrics}
We employ three datasets CIFAR-10, CIFAR-100 and Mini-ImageNet, which are very popular for OCIL. CIFAR-10 \cite{CIFAR} is split into 5 disjoint tasks with 2 classes per task. CIFAR-100 \cite{CIFAR} is split into 10 disjoint tasks with 10 classes per task. For Mini-ImageNet \cite{MiniImageNet}, we first merge the original three parts into one dataset. Then, it is split into 10 tasks, each of which contains 10 classes. The way for splitting datasets is the same as previous works \cite{mir,aser,scr,dvc}. We split the classes of each dataset in order and keep them fixed in all experiments. 

We mainly use the average accuracy, forgetting and intransigence \cite{ewc++}, which are the three most widely used metrics for continual learning. Let $a_{i,j} \in [0,1]$ be the accuracy evaluated on the held-out test set of task $j$ after training the network from task 1 to $i$. $f_{i,j}$ represents how much the model forgets about task $j$ after being trained on task $i$. To reduce computation consumption, we use the definition in \cite{ave_intransi} for intransigence. In detail, $a_j^{*}$ denotes the accuracy evaluated on the held-out test set of task $j$ by a reference model, which is fine-tuned from task 1 to task $j$ in an incremental way using the cross-entropy loss.  For $T$ tasks, the \textbf{Average Accuracy}, \textbf{Average Forgetting} and \textbf{Average Intransigence} could be formulated as:
\begin{equation}
	Average \  Accuracy(A_{T}) = \frac{1}{T} \sum_{j=1}^{T} a_{T,j}.
\end{equation}
\begin{align}
	Average \  Forgetting(F_{T}) &= \frac{1}{T-1} \sum_{j=1}^{T-1} f_{T,j}, \\
	\text{where} \  f_{i,j} &= \max \limits_{l \in \{1,\ldots,i-1 \} } a_{l,j} - a_{i,j}.&& \notag
\end{align}
\begin{align}
	Average \  Intransigence (I_{T}) &= \frac{1}{T} \sum_{j=1}^{T} a_j^{*} - a_{j,j}.
\end{align}
Therefore, the value range of these metrics is: $A_{T}\in[0,1]$, $F_{T}\in[-1,1]$ and $I_{T}\in[-1,1]$. And note that the lower $F_{T}, I_{T}$, the higher stability (lower forgetting) and higher plasticity (better forward transfer) of a method. Besides, to measure the overall feasibility, we also report the total running time including training and testing time at the same experiment environment. All scores except the total running time are the average of 10 runs.

\begin{table*}[ht]
	\centering
	\caption{Comparison of Average Accuracy ($\%$) $\pm$ Confidence Interval ($t_{0.975}$) on over 10 runs. Higher is better ($\uparrow$). Note that SCR has adopted the review trick acquiescently in its official code.}
	\scalebox{0.75}{
		\begin{tabular}{cccc|ccc|ccc}
			\toprule
			Method & & Mini-ImageNet & & & CIFAR-100 & & & CIFAR-10 & \\
			
			\midrule
			Memory & M=1k & M=2k & M=5k & M=1k & M=2k & M=5k & M=0.2k & M=0.5k & M=1k \\
			
			\midrule\midrule
			fine-tune & \multicolumn{3}{c}{$3.9 \pm 0.6$} & \multicolumn{3}{c}{$5.7 \pm 0.2$} & \multicolumn{3}{c}{$17.7 \pm 0.4$} \\
			
			iid offline & \multicolumn{3}{c}{$51.4 \pm 0.2$} & \multicolumn{3}{c}{$49.6 \pm 0.2$} & \multicolumn{3}{c}{$81.7 \pm 0.1$} \\
			
			\midrule
			AGEM (ICLR’19) & $4.2 \pm 0.4$ & $4.4 \pm 0.3$ & $4.4 \pm 0.2$ & $5.9 \pm 0.2$ & $6.0 \pm 0.3$ & $5.8 \pm 0.3$ & $18.0 \pm 0.5$ & $18.2 \pm 0.2$ & $18.3 \pm 0.2$ \\
			AGEM-RV & $7.0 \pm 0.8$ & $11.6 \pm 1.1$ & $18.4 \pm 0.8$ & $6.6 \pm 0.3$ & $9.4 \pm 1.3$ & $15.4 \pm 1.3$ & $18.2 \pm 0.3$ & $19.7 \pm 1.8$ & $26.4 \pm 2.7$ \\
			\midrule
			
			ER (ICML-W’19) & $10.6 \pm 0.6$ & $12.8 \pm 1.1$ & $15.4 \pm 1.1$ & $11.8 \pm 0.4$ & $14.5 \pm 0.9$ & $20.1 \pm 1.0$ & $23.2 \pm 1.0$ & $31.3 \pm 1.7$ & $36.7 \pm 2.9$ \\
			ER-RV & $14.8 \pm 0.7$ & $21.7 \pm 0.4$ & $27.9 \pm 0.6$ & $14.5 \pm 0.4$ & $21.2 \pm 0.4$ & $30.9 \pm 0.6$ & $26.2 \pm 1.6$ & $39.2 \pm 1.9$ & $50.0 \pm 1.7$ \\
			\midrule
			
			GSS (NeurIPS’19) & $10.2 \pm 0.7$ & $12.9 \pm 1.2$ & $15.4 \pm 1.1$ & $10.1 \pm 0.5$ & $13.7 \pm 0.6$ & $17.4 \pm 0.9$ & $22.8 \pm 1.3$ & $29.2 \pm 1.2$ & $35.3 \pm 2.5$ \\
			GSS-RV & $13.7 \pm 0.3$ & $19.9 \pm 0.9$ & $26.3 \pm 0.7$ & $12.8 \pm 0.4$ & $18.0 \pm 0.3$ & $25.8 \pm 0.9$ & $23.7 \pm 0.8$ & $34.6 \pm 1.4$ & $45.7 \pm 1.5$ \\
			\midrule
			
			MIR (NeurIPS’19) & $10.4 \pm 0.7$ & $15.4 \pm 1.1$ & $18.7 \pm 1.1$ & $11.0 \pm 0.4$ & $15.5 \pm 0.6$ & $22.1 \pm 0.8$ & $23.8 \pm 1.2$ & $30.7 \pm 2.4$ & $42.8 \pm 1.5$ \\
			MIR-RV & $13.1 \pm 0.9$ & $19.7 \pm 0.7$ & $28.2 \pm 0.6$ & $12.9 \pm 0.5$ & $18.7 \pm 0.4$ & $30.5 \pm 0.8$ & $25.2 \pm 2.2$ & $35.8 \pm 1.7$ & $46.5 \pm 1.5$ \\
			\midrule
			
			
			GDumb (ECCV’20) & $8.1 \pm 0.4$ & $12.4 \pm 0.6$ & $20.5 \pm 0.6$ & $10.1 \pm 0.3$ & $14.4 \pm 0.3$ & $20.9 \pm 0.3$ & $27.6 \pm 1.3$ & $33.1 \pm 1.2$ & $38.8 \pm 1.1$ \\
			\midrule
			
			ASER (AAAI’21) & $12.4 \pm 0.8$ & $14.4 \pm 1.0$ & $16.3 \pm 2.3$ & $14.9 \pm 0.6$ & $18.8 \pm 0.7$ & $23.5 \pm 0.7$ & $30.3 \pm 1.4$ & $39.4 \pm 1.6$ & $46.6 \pm 1.4$ \\
			ASER-RV & $17.0 \pm 0.4$ & $20.3 \pm 0.5$ & $25.1 \pm 0.8$ & $18.1 \pm 0.3$ & $23.9 \pm 0.7$ & $31.0 \pm 0.6$ & $26.1 \pm 0.8$ & $34.6 \pm 1.0$ & $44.8 \pm 1.8$ \\
			\midrule
			
			ER-ACE (ICLR’22) & $17.0 \pm 0.8$ & $19.1 \pm 0.7$ & $20.7 \pm 0.8$ & $20.1 \pm 0.7$ & $21.8 \pm 1.1$ & $22.8 \pm 0.9$ & $43.8 \pm 1.4$ & $50.5 \pm 1.4$ & $52.2 \pm 1.6$ \\
			ER-ACE-RV & $18.8 \pm 0.5$ & $22.1 \pm 0.5$ & $25.1 \pm 0.4$ & $21.4 \pm 0.6$ & $24.7 \pm 0.5$ & $26.6 \pm 0.7$ & $45.9 \pm 1.6$ & $54.7 \pm 1.6$ & $59.3 \pm 1.5$ \\
			\midrule
			
			DVC (CVPR’22) & $15.5 \pm 1.4$ & $18.1 \pm 1.4$ & $19.2 \pm 1.5$ & $20.6 \pm 0.5$ & $22.6 \pm 0.7$ & $25.2 \pm 1.1$ & $44.5 \pm 2.8$ & $50.2 \pm 3.2$ & $51.6 \pm 2.3$ \\
			DVC-RV & $20.7 \pm 1.2$ & $24.0 \pm 1.4$ & $27.4 \pm 1.5$ & $26.1 \pm 0.8$ & $30.7 \pm 0.7$ & $35.9 \pm 1.0$ & $49.8 \pm 1.5$ & $56.1 \pm 2.4$ & $59.4 \pm 1.2$ \\
			\midrule
			
			
			AAER (TCSVT’22) & $23.0 \pm 1.3$ & $25.2 \pm 1.6$ & $26.6 \pm 1.6$ & \uline{$27.5 \pm 0.8$} & $ 30.3 \pm 0.5$ & $33.1 \pm 0.6$ & $50.5 \pm 0.6$ & \uline{$61.4 \pm 0.9$} & $63.7 \pm 1.9$ \\
			\midrule
			
			OCM (ICML’22) & $12.4 \pm 0.8$ & $13.3 \pm 1.0$ & $14.4 \pm 0.9$ & $20.6 \pm 1.0$ & $25.6 \pm 0.8$ & $29.5 \pm 1.2$ & \uline{$53.0 \pm 3.0$} & $59.5 \pm 1.7$ & $65.2 \pm 1.6$ \\
			\midrule
			
			SCR (CVPR-W’21) & \uline{$25.0 \pm 0.4$} & \uline{$30.4 \pm 0.6$} & \uline{$34.6 \pm 0.6$} & $27.1 \pm 0.5$ & \uline{$32.8 \pm 0.4$} & \uline{$37.7 \pm 0.6$} & $49.8 \pm 1.1$ & $60.4 \pm 0.9$ & \uline{$66.7 \pm 0.6$} \\
			\midrule
			
			AFS (Ours) & \bm{$26.0 \pm 0.5$} & \bm{$31.7 \pm 0.6$} & \bm{$37.6 \pm 0.3$} & \bm{$27.8 \pm 0.7$} & \bm{$35.0 \pm 0.6$} & \bm{$42.3 \pm 0.4$} & \bm{$55.9 \pm 1.5$} & \bm{$63.4 \pm 0.6$} & \bm{$67.9 \pm 0.5$} \\ 
			\midrule
			\bottomrule
			
		\end{tabular}
	}
	
	\label{average_accuracy}
\end{table*}

\subsection{Implementation Detail}
\subsubsection{Setting}
For fair comparison, the setting is the same as \cite{agem,mir,aser,der,scr}. We adopt a reduced ResNet-18 \cite{resnet} as the base model for all datasets, which is optimized using SGD. The learning rate is set to 0.1. The batch size for samples from data stream is 10 while that for the batch retrieved from memory is 100. Like ER \cite{er}, we adopt the reservoir memory update and random retrieval strategy. 

\subsubsection{Hyper-parameters Selection}
The hyper-parameters of AFS are $\alpha=0.25$, $\sigma=0.5$, $\mu=0.3$, $T=20$, $\epsilon=0.01$ for all experiments. For $\beta$, we think different memory size needs different strength of regularization by VKD. Thus, we search the best $\beta$ from $[0.01,0.05,0.1,0.15,0.2,0.25,0.3,0.4,0.5]$ for different memory sizes and datasets by the method in \cite{mai_online_survey}. When the temperature is 1, the VKD degrades to LSR. 

\subsubsection{Data Augmentation}
Since data augmentation is a common trick to improve performance, we also apply it to AFS, which can further help RFL to find real ASI samples. Specifically, we utilize the random-resized-crop, random-horizontal-flip and color-jitter for all datasets. An additional random-gray-scale is used for CIFAR-10. Note that we only use data augmentation for batches retrieved from memory. 

\subsubsection{Review Trick}
Following previous implementations \cite{eeil,ber,scr}, we adopt the review trick (RV), which is a 1-epoch fine-tuning stage with a small learning rate (i.e. $0.01$) using all samples in the memory buffer. Specifically, it is used after training a new task or before an inference time. When the task boundary is not clear, we suggest the RV is started after every 500/1000 iterations. We do not use any data augmentation and regularization in RV (i.e., only RFL is used in RV). The batch size of RV is set to 10.

\begin{figure*}[ht]
	\centering 
	\subfloat{\includegraphics[scale=0.37]{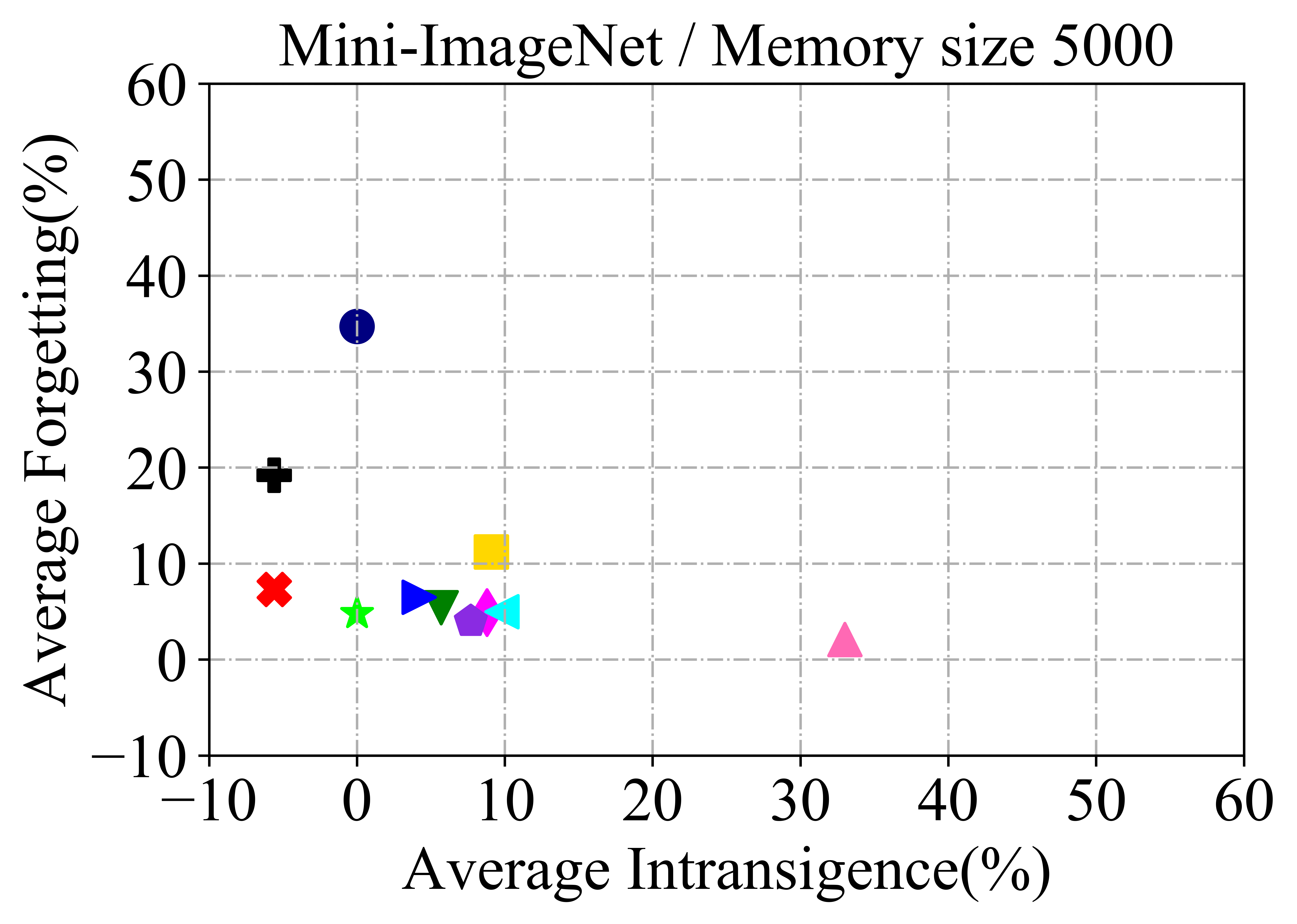}}
	\subfloat{\includegraphics[scale=0.37]{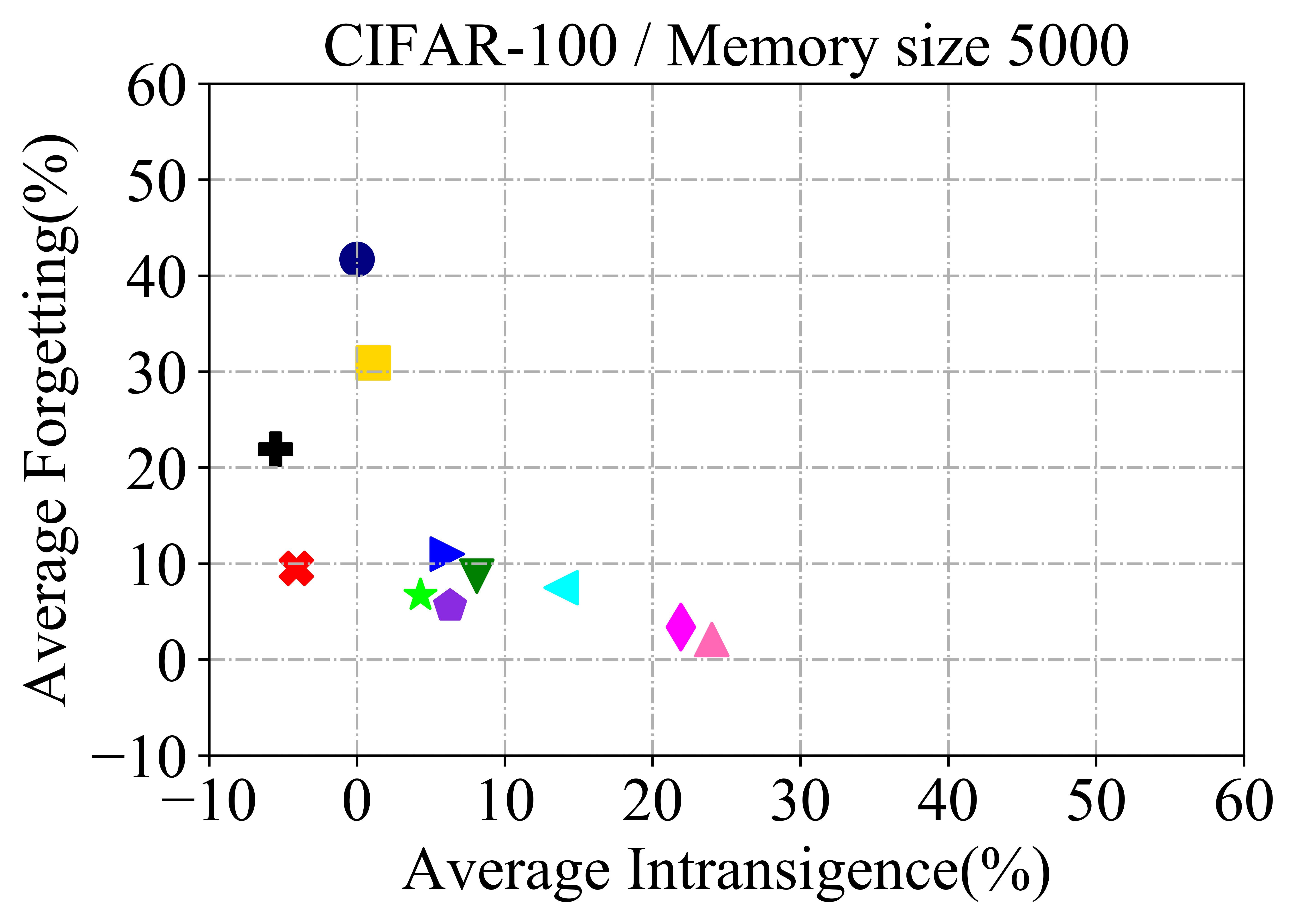}}
	\subfloat{\includegraphics[scale=0.37]{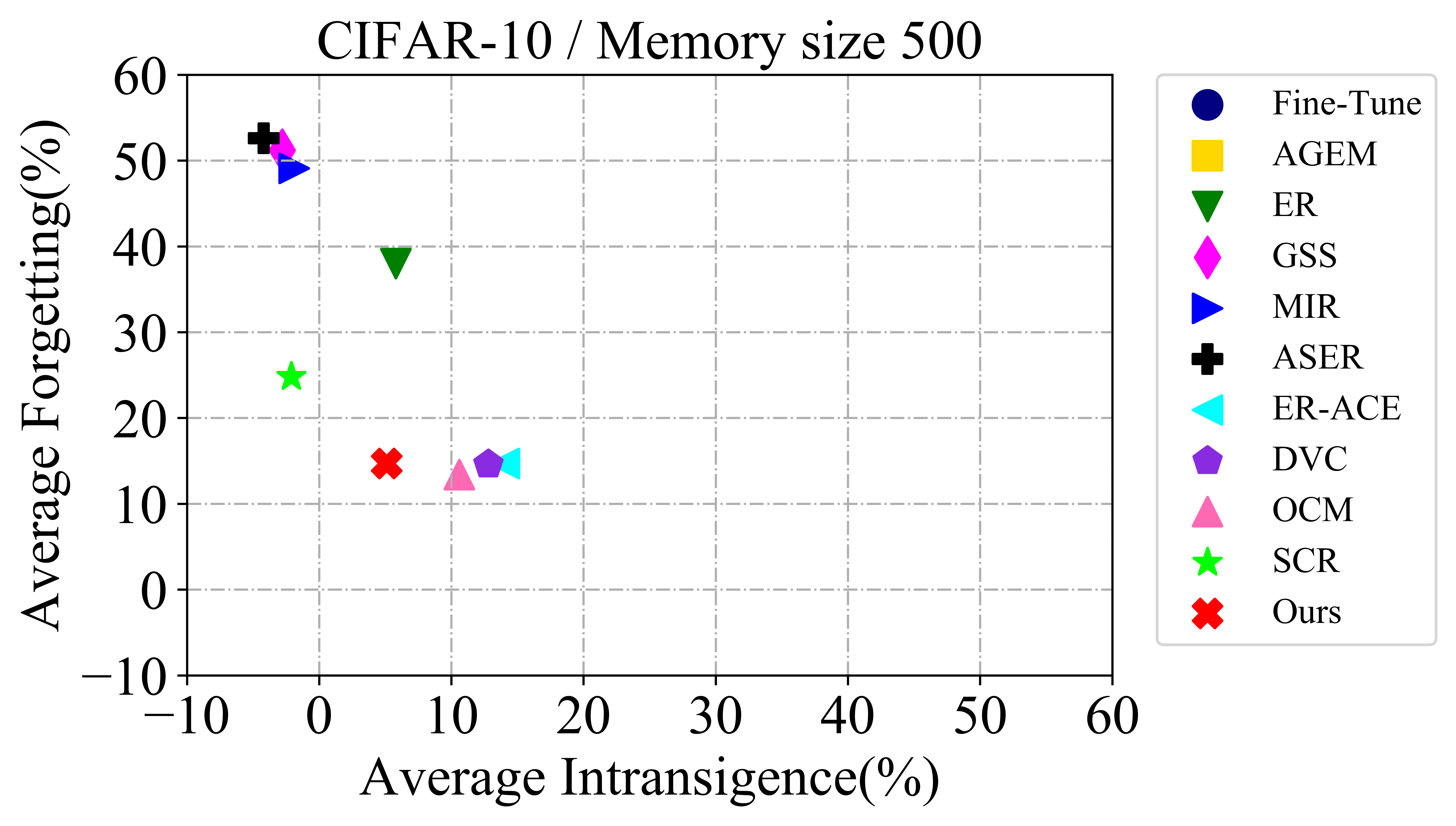}}
	\caption{Interplay between average intransigence and average forgetting. The average results are calculated over 10 runs.}
	\label{average_forgetting_and_intransigence}
\end{figure*}

\begin{figure*}[ht]
	\centering 
	\subfloat{\includegraphics[scale=0.37]{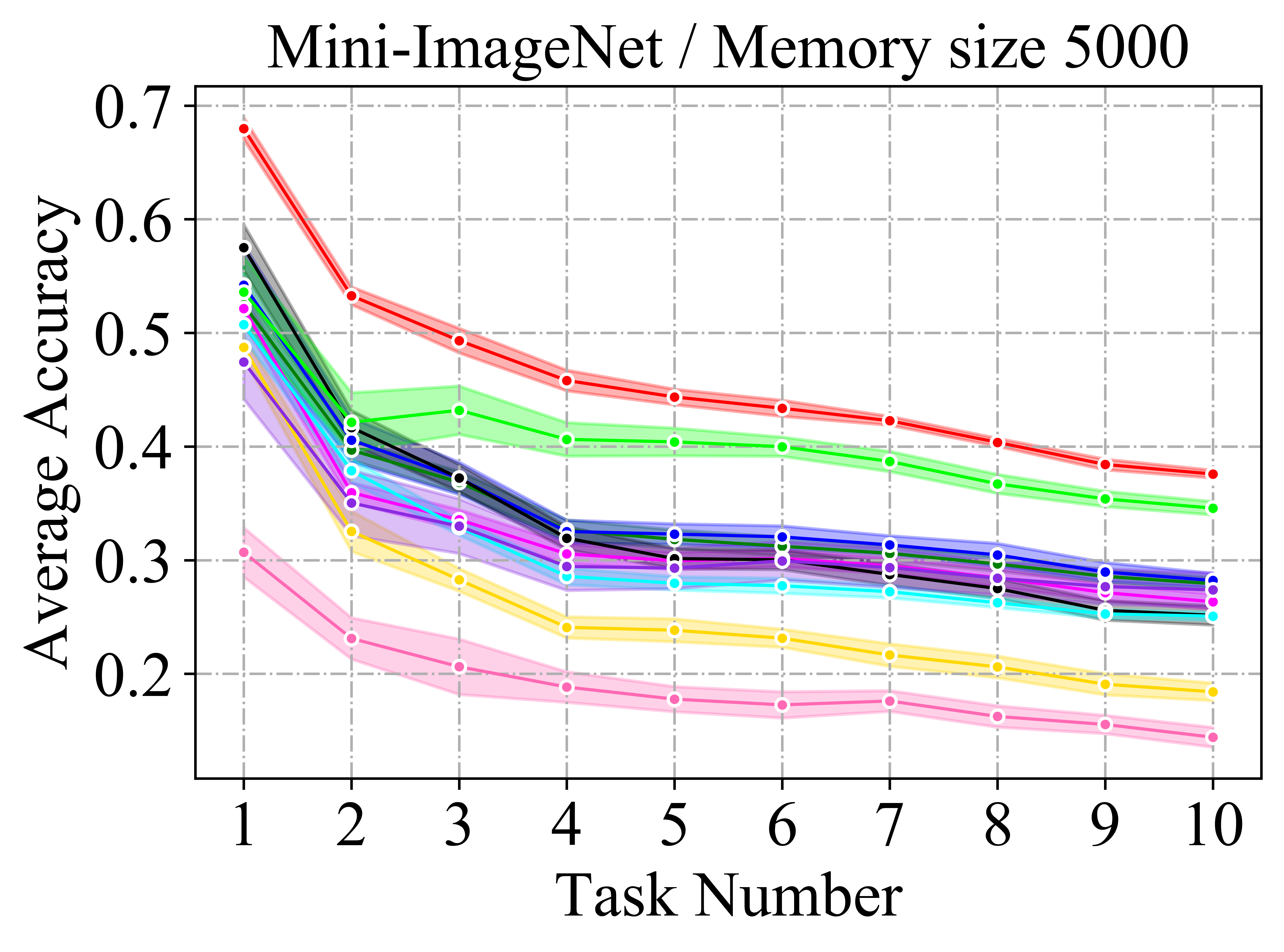}}
	\subfloat{\includegraphics[scale=0.37]{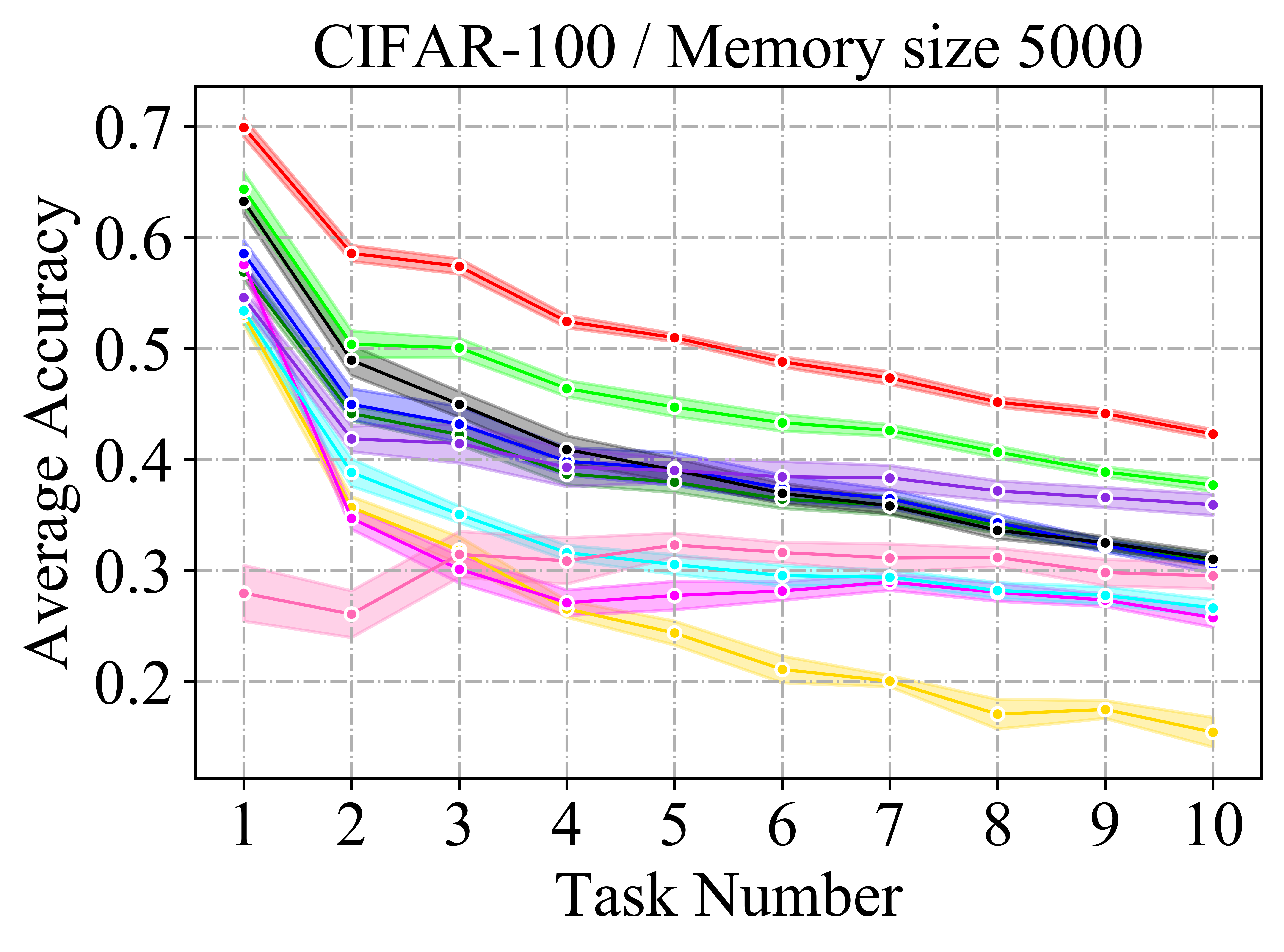}}
	\subfloat{\includegraphics[scale=0.37]{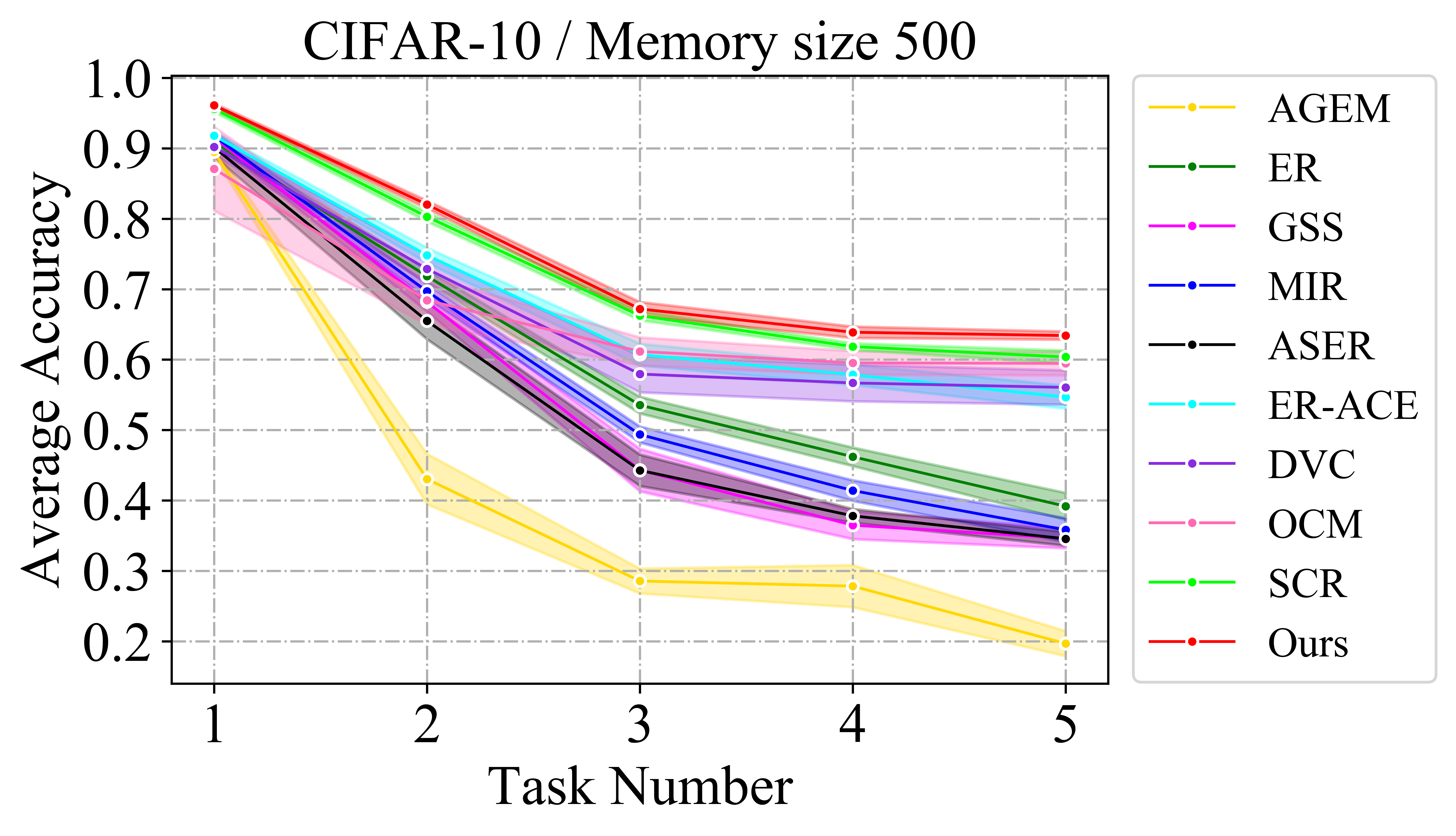}}
	\caption{Average Accuracy $\pm$ Confidence Interval ($t_{0.975}$) after each incremental step. All baselines adopt RV except OCM.}
	\label{average_incremental_result}
\end{figure*}

\begin{figure}[t]
	\centering 
	\subfloat[Running time comparison.]{\includegraphics[height=1.1in, width=1.46in]{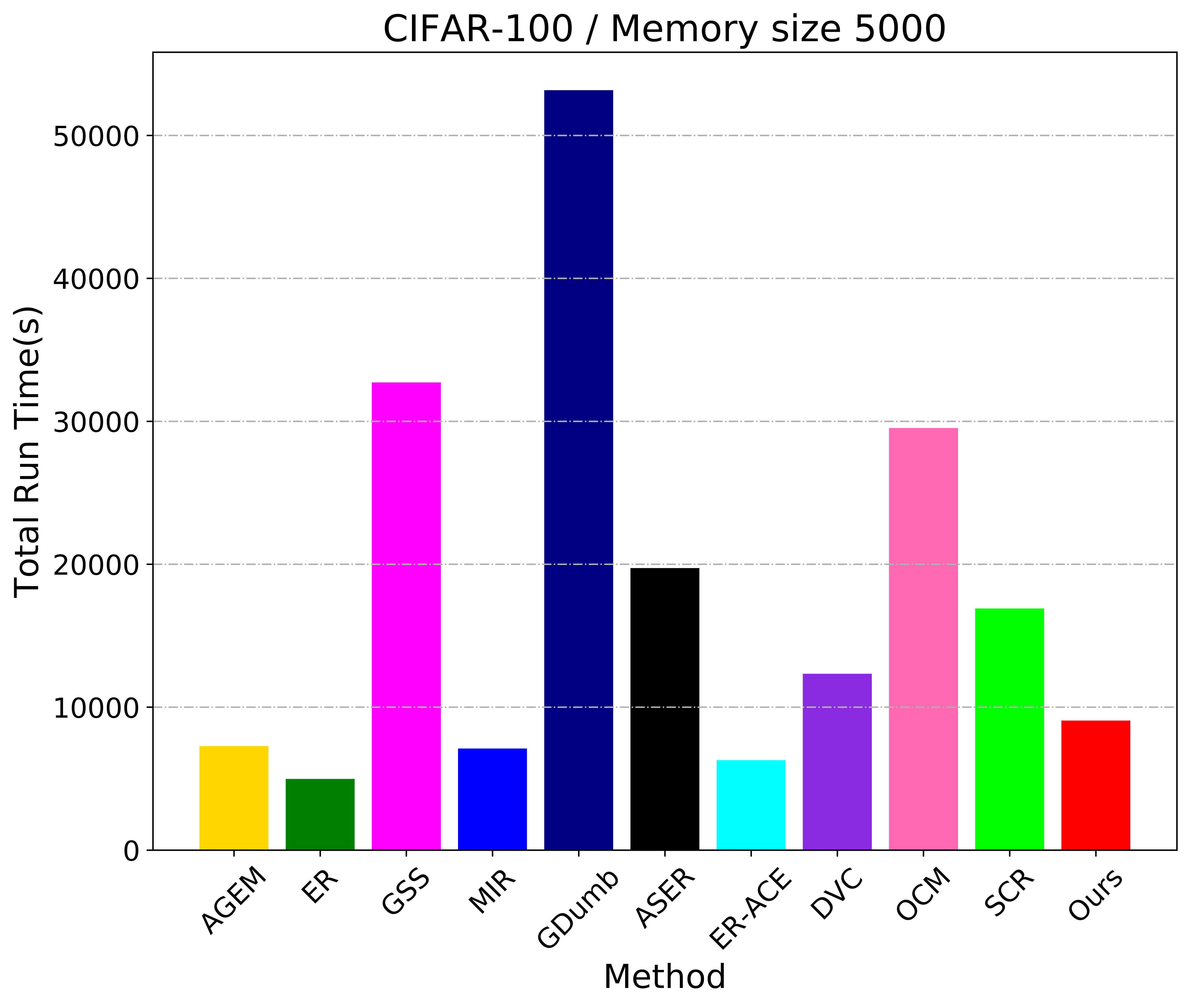} 
		\label{run_time}}
	\subfloat[Impact of retrieved batch size.]{\includegraphics[height=1.1in, width=1.5in]{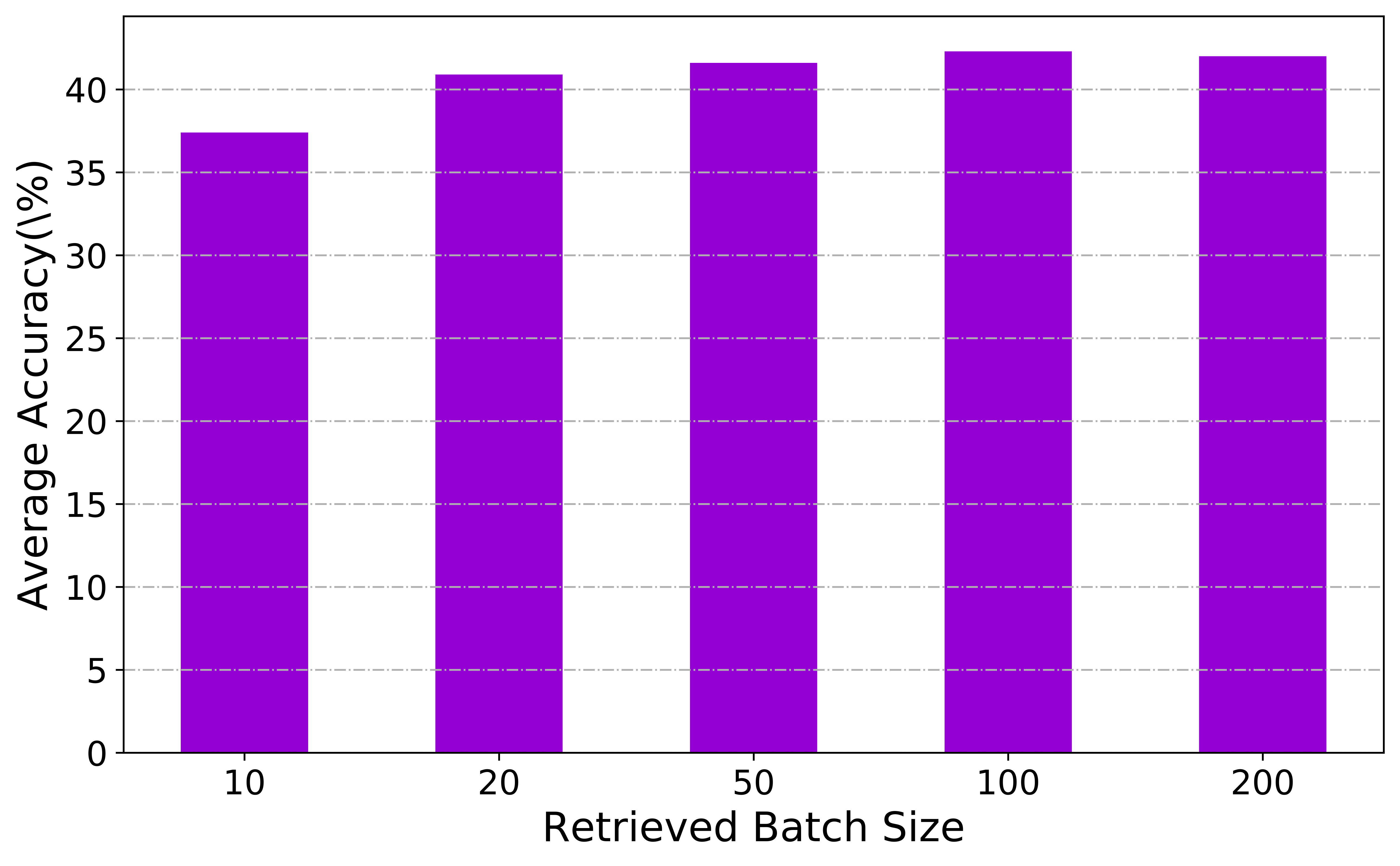}
		\label{impact_of_memory_batch}}
	\caption{Comparison of running time and influence of batch size on CIFAR-100 with 5k memory.}
	\label{run_time_and_batch_size}
\end{figure}

\subsection{Comparison with State-of-The-Art Methods}
We compare our AFS with several state-of-the-art algorithms for OCIL, which include \textbf{AGEM} \cite{agem}, \textbf{ER} \cite{er}, \textbf{GSS} \cite{gss}, \textbf{MIR} \cite{mir}, \textbf{GDumb} \cite{gdumb}, \textbf{ASER} \cite{aser}, \textbf{ER-ACE} \cite{er_ace}, \textbf{DVC} \cite{dvc}, \textbf{AAER} \cite{aaer}, \textbf{OCM} \cite{ocm} and \textbf{SCR} \cite{scr}. For most of these methods, we report the results using the official code and the optimal hyper-parameters stated in their papers/code. However, according to our experiments, OCM obtains unfair benefit from the Adam optimizer and ResNet-18, which is more powerful than the widely used SGD and reduced ResNet-18. Therefore, we change them to the traditional SGD and reduced ResNet-18 for fair comparison, where we search the best learning rate and weight decay. Since AFS employs the memory retrieval and update strategy of ER, ER and its variation are our baselines. Besides, we will also list the performance of \textbf{fine-tune} and \textbf{iid-offline}, which are regarded as the lower and upper bound for OCIL respectively. 

We first compare our model to these newly published methods. Table \ref{average_accuracy} shows the average accuracy at the end of data stream on three datasets over 10 runs. From this table, we can see that AFS outperforms all methods on all three datasets under any settings. And in most cases, the margin becomes larger when the memory size keeps rising. For example, compared to the current best score, the margin increases from $1.0\%$ to $3.0\%$ on the Mini-ImageNet, from $0.3\%$ to $4.6\%$ on the CIFAR-100. In contrast, the margin on CIFAR-10 becomes smaller. We think this is because CIFAR-10 is an easy dataset and the performance of reduced ResNet-18 tends to saturate. Besides, when adopting the reduced ResNet and SGD optimizer, the performance of OCM becomes a little worse, which shows the unfair benefit of better backbone and optimizer. To relieve catastrophic forgetting, SCR replaced the soft-max and cross-entropy with Nearest-Class-Mean classifier \cite{scr}. Although it shows better performance, it is more difficult for this classifier to rightly predict the outliers. In contrast, we propose new insight by modifying CE to tackle this problem.

According to our experiments, RV can improve the performance for most methods, which is also pointed in \cite{mai_online_survey}. Therefore, for fair comparison, we further add RV to all baselines except GDumb, OCM and AAER. GDumb uses all memory samples to train a model from scratch. Due to the huge consumption of OCM and the unavailability of official code for AAER, we do not apply RV for them. While our method does not use data augmentation in RV, ER-ACE, DVC and SCR have adopted the combination of RV and data augmentation since these methods can benefit from them. Note that SCR has adopted RV actually in its official code. The corresponding results are labelled as '-RV'. As shown in Table \ref{average_accuracy}, RV can improve the accuracy in most cases. Only the ASER-RV fails on CIFAR-10. Through deep analysis on both the original algorithm and experiment results, we find the memory buffer contains more samples of new classes when using the ASER updating strategy. Therefore, using RV in ASER will aggravate task-recency bias. Compared with methods augmented with RV, our approach still outperforms them by a large margin in most cases. The margin on CIFAR-100 is a little smaller when the memory size is 1k. We attribute it to the more classes and limited resolution of images, which increases the difficulty of a model to learn more discriminative features.

Moreover, according to Table \ref{average_accuracy}, using random method for memory updating and retrieval obtains larger margin than some specially designed strategies when RV is added. In detail, ER adopts a random updating and retrieval method while GSS, MIR and ASER focus on designing better strategies. However, ER-RV outperforms GSS-RV, MIR-RV in most cases. And sometimes it even obtains higher accuracy than the more recent approach AESR-RV. Therefore, we take ER as our basic model and pay more attention to learning strategy. 

Since continual learning aims to strike a balance between stability and plasticity, hence it is more reasonable to consider the forgetting and intransigence together rather than individually. Fig. \ref{average_forgetting_and_intransigence} shows the interplay between average intransigence and average forgetting for each method. According to \cite{ewc++}, \textbf{the closer to the left lower corner, the better balance is struck.}  As shown in this figure, AFS is the closest to the left lower corner, which means that AFS strikes a best balance between stability and plasticity. Also, from this figure, we can see that even though some methods (e.g., GSS, DVC and OCM) obtain excellent average forgetting (higher stability), they show too high average intransigence (lower plasticity).

Fig. \ref{average_incremental_result} shows the average accuracy after each incremental step. In this figure, AFS consistently outperforms all methods, which illustrates that AFS is more effective. Furthermore, Fig. \ref{run_time_and_batch_size}\subref{run_time} gives the total running time (training + inference) on CIFAR-100 at 5k memory. RV is used when calculating the running time. It can be seen that AFS has achieved competitive speed with higher accuracy.

To show the distribution of samples according to model's prediction, we illustrate the number of three kinds of samples by ER-RV and our method in Fig. \ref{mean_weight_and_logit}\subref{sample_type}. \Rt{In this figure, compared with ER-RV (our baseline), our method consistently owns less HSI samples and more ESI samples. On the other hand, the number of ASI samples in our method is close to ER-RV in most cases. This means our method can promote HSI samples to be either ESI samples or ASI samples.} From this, we can conclude that our method actually learns the ASI samples more adequately by promoting their weights. And hence our method really can make the type of samples polarize by paying more attention to ASI sample (i.e. samples belong to either HSI or ESI, instead of locating at ASI).



The above results validate the effectiveness of sufficiently learning on ASI samples for relieving forgetting and improving plasticity.


\begin{table*}[t]
	\centering
	\caption{Influence of each module using three metrics ($\%$) over 10 runs. "AA", "AF" and "AI" refer to the Average Accuracy, Average Forgetting and Average Intransigence respectively. Note that the Baseline  contains data augmentation and larger batch size compared to ER-RV.}
	\scalebox{0.75}{
		\begin{tabular}{cccc|ccc|ccc}
			\toprule
			Method &  & Mini-ImageNet &  &   & CIFAR-100 &   &   & CIFAR-10 &  \\
			\midrule
			Memory & \multicolumn{3}{c}{M=5k} & \multicolumn{3}{c}{M=5k} & \multicolumn{3}{c}{M=0.5k} \\
			\midrule
			Metrics & AA$\uparrow$ & AF$\downarrow$ & AI$\downarrow$ & AA$\uparrow$ & AF$\downarrow$ & AI$\downarrow$ & AA$\uparrow$ & AF$\downarrow$ & AI$\downarrow$ \\
			\midrule
			\midrule
			
			ER-RV & $27.9 \pm 0.6$ & $5.4 \pm 0.3$ & $5.7 \pm 1.6$ & $30.9 \pm 0.6$ & $8.7 \pm 0.4$ & $8.1 \pm 0.8$ & $39.2 \pm 1.9$ & $38.0 \pm 2.3$ & $5.8 \pm 2.4$ \\
			
			Baseline & $31.0 \pm 1.0$ & $6.9 \pm 0.9$ & $2.2 \pm 2.6$ & $36.7 \pm 1.7$ & $8.0 \pm 0.6$ & $4.9 \pm 1.8$ & $56.3 \pm 1.4$ & $18.3 \pm 1.5$ & $8.6 \pm 1.9$ \\
			\midrule
					
			Baseline + FL & $35.5 \pm 0.6$ & $7.9 \pm 0.7$ & $-3.8 \pm 1.8$ & $41.1 \pm 0.2$ & $9.4 \pm 0.5$ & $-2.0 \pm 1.3$ & $57.6 \pm 0.9$ & $15.2 \pm 2.4$ & $11.0 \pm 2.8$ \\
			
			Baseline + RFL & $36.8 \pm 0.5$ & $7.7 \pm 0.4$ & $-4.5 \pm 1.7$ & $41.0 \pm 0.4$ & $9.5 \pm 0.6$ & $-2.8 \pm 1.2$ & $61.9 \pm 0.7$ & $14.3 \pm 1.3$ & $7.4 \pm 2.2$ \\
			\midrule
			
			Baseline + LSR & $32.0 \pm 1.2$ & $7.8 \pm 0.7$ & $-0.4 \pm 2.1$ & $38.4 \pm 1.6$ & $9.2 \pm 0.8$ & $1.1 \pm 1.4$ & $56.8 \pm 0.8$ & $18.3 \pm 1.1$ & $7.9 \pm 2.9$ \\
			
			Baseline + VKD & $32.3 \pm 1.2$ & $6.1 \pm 0.4$ & $1.4 \pm 1.9$ & $38.1 \pm 1.3$ & $7.8 \pm 0.7$ & $3.4 \pm 1.7$ & $58.4 \pm 1.7$ & $18.3 \pm 1.8$ & $6.3 \pm 1.9$ \\
			\midrule
			
			Baseline + FL  + LSR & $35.3 \pm 0.6$  & $8.6 \pm 0.3$ & $-4.4 \pm 1.9$ & $40.8 \pm 0.3$ & $9.6 \pm 0.5$ & $-1.9 \pm 1.0$ & $59.0 \pm 0.6$ & $15.1 \pm 1.7$ & $9.4 \pm 2.5$ \\
			
			Baseline + FL  + VKD & $36.6 \pm 0.7$ & $8.3 \pm 0.7$ & $-5.5 \pm 2.2$ & \bm{$42.4 \pm 0.2$} & $10.1 \pm 0.5$ & $-4.9 \pm 1.3$ & $60.9 \pm 1.2$  & $17.2 \pm 1.1$ & $5.0 \pm 2.0$ \\
			
			Baseline + RFL  + LSR & \uline{$36.9 \pm 0.5$} & $7.4 \pm 0.6$ & $-4.6 \pm 1.8$ & $41.2 \pm 0.4$ &  $8.6 \pm 0.6$ & $-2.0 \pm 1.6$ &  \uline{$62.3 \pm 0.7$} & $14.7 \pm 1.0$ & $6.5 \pm 1.9$ \\
			
			Baseline + RFL  + VKD & \bm{$37.6 \pm 0.3$} & $7.3 \pm 0.4$ & $-5.6 \pm 1.8$ & \uline{$42.3 \pm 0.4$} & $9.5 \pm 0.4$ & $-4.1 \pm 1.1$ & \bm{$63.4 \pm 0.6$} & $14.7 \pm 1.1$ & $5.1 \pm 2.1$ \\
			
			\midrule
			\bottomrule
		\end{tabular}
	}
	
	\label{ablation_study}
\end{table*}

\begin{table*}[t]
	\centering
	\caption{\Rt{Average accuracy of different methods for calculating the sample's weight.}}
	\scalebox{0.75}{
		\begin{tabular}{cccc|ccc|ccc}
			\toprule
			Method & & Mini-ImageNet & & & CIFAR-100 & & & CIFAR-10 & \\			
			\midrule
			Memory & M=1k & M=2k & M=5k & M=1k & M=2k & M=5k & M=0.2k & M=0.5k & M=1k \\
			\midrule
			\midrule
			
			Ours with SMU & $22.2 \pm 1.1$ & $27.9 \pm 1.0$ & $32.1 \pm 0.6$ & $25.2 \pm 0.8$ & $31.8 \pm 0.4$ & $39.8 \pm 0.6$ & $53.7 \pm 1.6$ & $62.2 \pm 0.7$ & $67.3 \pm 0.4$ \\

			Ours with GI & \uline{$24.6 \pm 0.5$} & \uline{$30.9 \pm 0.5$} & \uline{$36.1 \pm 0.5$} & \uline{$26.9 \pm 0.7$} & \uline{$33.9 \pm 0.4$} & \uline{$42.1 \pm 0.4$} & $54.0 \pm 2.6$ & \uline{$62.4 \pm 1.0$} & \uline{$67.4 \pm 0.4$} \\

			Ours with Entropy & \uline{$24.6 \pm 0.6$} & $28.8 \pm 0.8$ & $33.7 \pm 0.6$ & $25.0 \pm 0.7$ & $32.1 \pm 0.6$ & $38.8 \pm 0.5$ & \uline{$55.0 \pm 1.1$} & $62.3 \pm 0.7$ & $66.5 \pm 0.4$ \\

			Ours with GEN \cite{ash2019deep} & $20.4 \pm 0.7$ & $27.5 \pm 0.3$ & $33.8 \pm 0.6$ & $21.2 \pm 0.6$ & $27.2 \pm 0.5$ & $32.8 \pm 0.4$ & $46.0 \pm 1.0$ & $55.2 \pm 0.8$ & $61.6 \pm 1.2$ \\
			\midrule			
			
			Ours & \bm{$26.0 \pm 0.5$} & \bm{$31.7 \pm 0.6$} & \bm{$37.6 \pm 0.3$} & \bm{$27.8 \pm 0.7$} & \bm{$35.0 \pm 0.6$} & \bm{$42.3 \pm 0.4$} & \bm{$55.9 \pm 1.5$} & \bm{$63.4 \pm 0.6$} & \bm{$67.9 \pm 0.5$} \\ 
			\midrule
			\bottomrule
			
		\end{tabular}
	}
	\label{ablation_study_in_AL}
\end{table*}

\subsection{Ablation Study}
To validate the effectiveness of RFL and VKD, the two main components of AFS, we have conducted some ablation experiments on three datasets, whose results are shown in Table \ref{ablation_study}. In this table, the baseline is a modified ER. Compared with ER, this baseline adopts larger retrieved batch (100), additional data augmentation and RV. 

First, we compare the performance gain between FL and RFL. As shown in Table \ref{ablation_study}, both of them can improve the performance, but the margin of RFL is much larger in most cases. For example, on CIFAR-10, FL shows little improvement while the accuracy gain of RFL reaches $5.6\%$. On Mini-ImageNet, the accuracy improvement of FL achieves $4.5\%$ while RFL reaches $5.8\%$. On CIFAR-100, RFL and FL obtain similar performance. We hypothesize that, to some extent, the CIFAR-100 has been learned well under limited resolution of images. 

Second, we investigate the influence of VKD and compare it with label smoothing regularization (LSR). In Table \ref{ablation_study}, VKD performs a little better than LSR in most cases and the improvement becomes bigger on easy dataset. However, when they are combined with FL/RFL, VKD gets much better performance than LSR on all datasets. For instance, when they are added into Baseline + RFL respectively, the accuracy increase of LSR is $0.1\%$ while VKD gets $0.8\%$ on Mini-ImageNet. The corresponding accuracy increase on CIFAR-10 and CIFAR-100 are 0.4\%, 1.5\% and 0.2\%, 1.3\% respectively. The performance comparison by combining LSR/VKD to Baseline + FL is similar. \Rt{Besides, considering average intransigence (i.e. plasticity), VKD shows a little bad plasticity on the two harder datasets. However, when combining VKD with RL/RFL (i.e. change the focus to the special type samples), they consistently show higher plasticity than combining LSR with FL/RFL.} The above validates that VKD is more effective when bigger weights are assigned to ambiguous samples. With high temperature, the VKD encourages model to pay more attention to non-target classes, which can enhance the ability of learning discriminative features.

\begin{figure*}[h]
	\centering
	\includegraphics[scale=0.28]{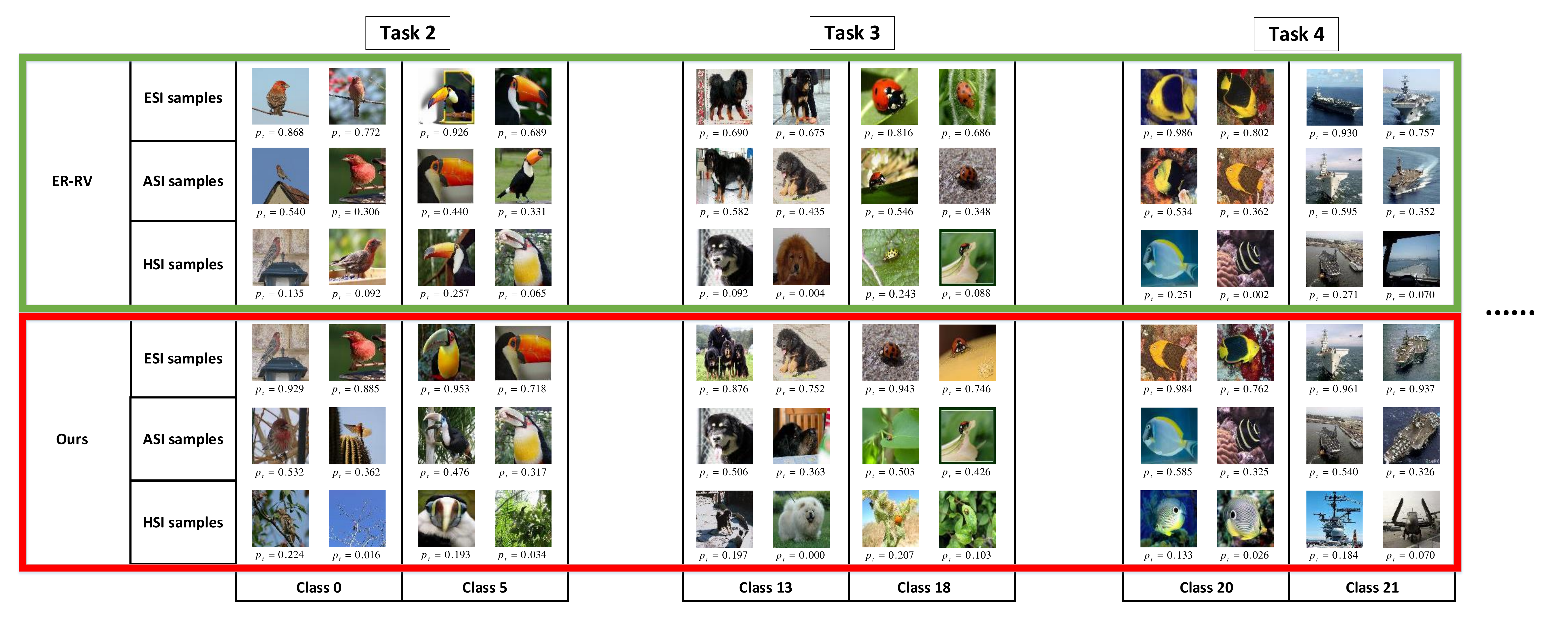}
	\caption{Sample images from Mini-ImageNet for three types of difficult samples, where the text below each image are the predicted possibility at target class by ER-RV and our AFS method. Note that the images are from two old classes in task $\mathcal{T}-1$ while current task id $\mathcal{T}$ is indicted in the top. We can find three type of samples in AFS are closer to human cognitive than ER-RV.}
	\label{visual_sample}
\end{figure*}

\Rt{Third, there exist some more complex methods in active learning (AL) \cite{2021deep_active} to measure the sample's value compared with our weight function only depending on the predicted scores. To explore their effect in incremental learning, we have conducted comparative experiments. Since most of the newest AL works focus on how to better use the unlabeled set, we finally select the "smallest margin uncertainty" (SMU), "gini-index" (GI), "Entropy" and "the norm of gradient embedding" (GEN \cite{ash2019deep}) to measure the uncertainty. For convenient comparison, we first normalize these values if they do not belong to $[0.0, 1.0]$. And we also apply the operation $1-U$ to $U_{GI}$, $U_{Entropy}$ and $U_{GEN}$. Thus, all these metrics satisfy the smaller value, the higher uncertainty. Next, we use these to replace our weight function. Note that we only use "Entropy" and "GEN" to define the weight and loss function does not include them. The resulting experimental results on three datasets are shown in Table \ref{ablation_study_in_AL}.}

\Rt{As shown in this Table, when adding these metrics into RFL, only GI achieves comparable performance when the memory size is large. For example, it achieves $67.4 \pm 0.4$ on CIFAR-10 at 1k memory and $42.1 \pm 0.4$ on CIFAR-100 at 5k memory. Other methods obtain much lower performance. In our opinion, the samples with higher ambiguity can also be the hard samples according to these new ambiguity metrics. Therefore, giving more attention to these hard samples will generate a tight classification boundary. This is benefit to active learning, where the class relationship is fixed and the model can be trained multiple times. However, in incremental learning, the class relationship is changed once new class data arrives. Thus, the tighter boundary is easier to be changed. Moreover, since the data can be used only once, the model is difficult to be converged. All these will lead to catastrophic forgetting. For example, the GEN, whose selected samples will change the model mostly, obtains the lowest performance in most cases.}


\begin{table}[t]
	\centering
	\caption{Average accuracy on CIFAR-100 using different $\alpha$ with 5k memory.}
	\scalebox{0.7}{
		\begin{tabular}{c|cccccc}
			\toprule
			$\alpha$ & $0.05$ & $0.1$ & $0.25$ & $0.5$ & $0.75$ & $1.0$ \\
			\midrule
			
			Ours & $25.3 \pm 0.4$ & $36.2 \pm 0.5$ & $42.3 \pm 0.4$ & $42.8 \pm 0.5$ & $40.4 \pm 1.7$ & $37.9 \pm 1.3$  \\
			\bottomrule
		\end{tabular}
	}
	\label{influence_of_alpha}
\end{table}

\begin{table}[t]
	\centering
	\caption{Average accuracy on CIFAR-100 using different $\sigma$ with 5k memory.}
	\scalebox{0.7}{
		\begin{tabular}{c|cccccc}
			\toprule
			$\sigma$ & $0.05$ & $0.1$ & $0.25$ & $0.5$ & $0.75$ & $1.0$ \\
			\midrule
			
			Ours & $8.3 \pm 0.3$ & $23.7 \pm 1.1$ & $41.0 \pm 0.4$ & $42.3 \pm 0.4$ & $42.4 \pm 0.4$ & $42.4 \pm 0.3$  \\
			\bottomrule
		\end{tabular}
	}
	\label{influence_of_sigma}
\end{table}

Lastly, we show the influence of some parameters. Table \ref{influence_of_alpha} gives the accuracy using difference $\alpha$ on CIFAR-100 with 5K memory. We can find that AFS reaches the highest scores when $\alpha$ is 0.25 or 0.5. For fair comparison with FL, we use 0.25 for $\alpha$ in other experiments. And Table \ref{influence_of_sigma} illustrates the accuracy on CIFAR-100 with different $\sigma$. We can see that the $\sigma$ should not be too small. This validates that easy samples should not be neglected absolutely as stated in previous sections. Finally, we validate the influence of retrieved batch size, whose results are given in Fig. \ref{run_time_and_batch_size}\subref{impact_of_memory_batch}. According to this figure, AFS achieves the best average accuracy when the memory batch size is 100.

\subsection{Visualization}
\Rt{To further give an intuitive understanding of three types of samples, we show some samples in Fig. \ref{visual_sample}, where the number below each image is the predicted probability at target class by ER-RV or our method after a specific task. Note that the images are from two old classes in task $\mathcal{T}-1$ while the current task id $\mathcal{T}$ is indicated at the top of this figure. From this figure, we can find the HSI samples in our method always contain more complex background or occluded objects while the objects in ESI samples are clear and easy to distinguish, which conforms with human perception. Even for the HSI samples, the predicted score by our method is not very small. For example, the 0.224 for the fifth image in class 0, whose background is very similar to the bird. In contrast, the HSI samples in ER-RV are much easier. Even in this case, it outputs much lower probability. Moreover, we can also find our predicted probability is much bigger than that of ER-RV. For example, the likelihood for the third image in Toucan (Class 5) by ER-RV is 0.331 while ours is 0.718. And similar contrast can also be found in other classes. This illustrates our model is more confident than ER-RV. Furthermore, the proposed method can grasp fined features for sub-class with less samples. For instance, for House Finch (Class 0), the character of male one is red feathers on its head/chest while the female only has taupe feathers. Since the dataset contains many male ones, intuitively, most of the male samples can be classified easily. However, from the Fig. \ref{visual_sample}, ER-RV regards some male House Finches with simple background as HSI samples. In contrast, we only regard the female one or images with complex background as HSI samples. In a word, through shift the focus to current ambiguous samples, we can learn more fined features and generate more confident probability.} 

\begin{table*}[ht]
	\centering
	\caption{Top-1 classification accuracy on ImageNet-LT for Open Long-Tailed Recognition. $\text{OLTR}^{\dag}$ represent the newly updated result in its website while OLTR is our implementation result using their code \protect\footnotemark.}
	\scalebox{0.85}{
		\begin{tabular}{c|cccc|cccc}
			\toprule
			
			Backbone Net &  \multicolumn{4}{c}{closed-set setting } \vline & \multicolumn{4}{c}{open-set setting } \\
			\midrule
			
			ResNet-10 & $>100$ & $\leqslant100 \ \& >20$ & $<20$ &  & $>100$ & $\leqslant100 \ \& >20$ & $<20$ &   \\
			\midrule
			
			Methods & Many-shot & Medium-shot & Few-shot & Overall & Many-shot & Medium-shot & Few-shot & F-measure  \\
			
			\midrule
			$\text{OLTR}^{\dag}$ & {$46.0$} & {$36.9$} & \uline{$18.4$} & {$37.8$} & {$44.2$} & {$35.2$} & \uline{$17.5$} & {$44.6$} \\
			
			OTLR & {$46.3$} & \bm{$37.9$} & \bm{$19.0$} & {$38.4$} & {$44.8$} & \bm{$36.5$} & \bm{$18.0$} & {$45.4$} \\
			
			\midrule
			
			OTLR + FL & {$48.9$} & {$36.1$} & {$14.5$} & {$37.9$} & \uline{$47.2$} & {$34.6$} & {$13.7$} & {$45.0$} \\
			
			OTLR + RFL & {$48.1$} & \uline{$37.8$} & \uline{$17.1$} & \uline{$38.8$} & {$46.6$} & \bm{$36.5$} & {$16.2$} & \uline{$45.9$} \\

			OTLR + $\text{RFL}^{d}$ & \bm{$49.8$} & {$36.8$} & {$17.0$} & \bm{$39.0$} & \bm{$48.2$} & \uline{$35.6$} & {$16.3$} & \bm{$46.1$} \\
			\bottomrule
			
		\end{tabular}
	}
	\label{oltr_imagenet}
\end{table*}
\footnotetext{\url{https://github.com/zhmiao/OpenLongTailRecognition-OLTR}} 

\subsection{Evaluation on Open Long-Tailed Recognition}
\textcolor{black}{
Except the CIL task, the class imbalance is also an important challenge in some other fields, such as open long-tailed recognition \cite{oltr}, dense long-tailed object detection \cite{eql,efl}. To further illustrate the effect of proposed RFL in these fields, we have done some exploratory experiments for the former task since it is more similar to CIL.}

\textcolor{black}{
For simplicity, we adopt the classical method \cite{oltr} (denoted as OLTR) as our baseline. The OLTR consists of dynamic meta-embedding and modulated attention. The authors employed a combination of cross-entropy classification loss and large-margin loss to optimize the parameters. Note that a two-stage training protocol is employed according to their code, which is a little different from their paper. Refer to the  paper \cite{oltr} and their code for more details. To validate the effect of RFL, we replace the cross-entropy classification loss with the proposed RFL loss in the loss function for the second stage. We follow the original setting and only adjust the hyper-parameters of the optimizer (e.g., learning rate and decay scheduler step) to make them more suitable. However, the performance gain of RFL is a little small in our initial experiments. We attribute this to the access of the distribution of whole classes and many training epochs in this task. To further improve the performance, we propose to dynamically calculate the $\mu_k$ for each class $k$, which is represented as $\text{RFL}^{d}$. In detail, for each class $k$, we split the predicted score into 25 bins to do the histogram statistics, where the score of each sample is obtained during the model’s forward process in the last epoch. Then, we can calculate a value $M_k$, which represents the difficulty of classifying this class
\begin{equation}
 \min{M_k} \quad s.t. \sum_{i=0}^{M_k} bin_{k, i} \geqslant cnt[k] * b 
\end{equation}
where $bin_{k, i}$ is number of samples located in $i$-th bin for class $k$, $cnt[k]$ is the total number of samples for class $k$. $b$ is a hype-parameter, which is set to $0.25$ through experiments. Finally, the $\mu$ in Eq. (2) is updated according to the following equation $\mu_k=M_k \times 0.04$. Through these two equations, we can dynamically adjust the focus according to the distribution of predicted scores. In other words, if the number of samples with small scores is larger, we should decrease the $\mu$ to focus more on harder samples. Otherwise, the hard samples is not very much, so we should attend to ambiguous samples. Note that we set all $\mu_k$ to 0.0 in the first epoch.} 

\textcolor{black}{
For convenience, we only conduct experiments on the ImageNet-LT dataset since the feature extraction network for this dataset is more similar to OCIL's. The experimental results are shown in Table \ref{oltr_imagenet}. From this table, we can see that ``OLTR + $\text{RFL}^{d}$'' achieves the best performance in both closed-set and open-set settings. Specifically, compared with ``OLTR'', it improves the overall performance by $0.6\%$ in closed-set setting and $0.7\%$ in open-set setting. Besides, the Top-1 accuracy improvement by adding he original "RFL" is larger than $0.4\%$ in both settings. In contrast, although using "FL" obtains better performance for many-shot classes, the accuracy of few-shot classes reduces significantly, which degrades the overall accuracy. We can conclude that it is important for few-shot and medium-shot classes to grasp both easy samples and ambiguous samples rather than hard samples. Finally, for many-shot classes, all the three loss functions get better performance, which means that easy samples should be down-weighted while more attention should be paid to ambiguous samples and hard samples. In a word, these experiments illustrate the effect of the proposed loss function in this task. Note that the "$\text{RFL}^{d}$" is not suitable for OCIL task since all samples for old tasks are not available.
}

\section{Conclusion}
In this paper, we propose a simple yet effective method AFS for online class incremental learning. After a detailed analysis of the task-recency bias caused by the class imbalance problem in OCIL, we found that ASI samples (potentially valuable samples) are critical to alleviate forgetting and improve plasticity. Therefore, we propose a RFL, which concentrates on these samples by shifting its focus. What is more, we introduce a VKD, which can further improve plasticity by avoiding over-fitting to target class. Extensive experiments on three popular datasets have shown the superiority of our method.

\normalem
\bibliographystyle{IEEEtran}
\bibliography{IEEEabrv,mypaper}


\begin{IEEEbiography}
[{\includegraphics[width=1in,height=1.25in,clip,keepaspectratio]{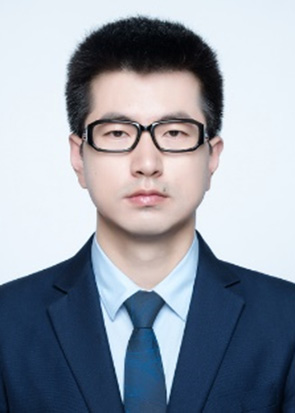}}]{Guoqiang Liang} received the B.S. in automation and the Ph.D. degrees in pattern recognition and intelligent systems from Xi’an Jiaotong University, Xi’an, China in 2012 and 2018 respectively. From Aug. 2018 to Aug 2020, he did the Post-Doctoral Research at the School of Computer Science, Northwestern Polytechnical University (NWPU), Xi’an, China. Currently, he is an associate professor at NWPU. His research interests include continual learning and video summarization.
\end{IEEEbiography}

\begin{IEEEbiography}
[{\includegraphics[width=1in,height=1.25in,clip,keepaspectratio]{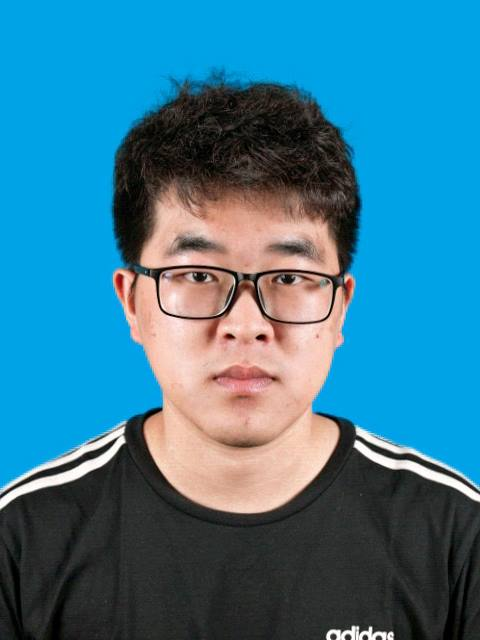}}]{Zhaojie Chen}
currently is a Master student in Northwestern Polytechnical University, Xi'an, China. His research interests include deep learning and continual learning.
\end{IEEEbiography}

\begin{IEEEbiography}
[{\includegraphics[width=1in,height=1.25in,clip,keepaspectratio]{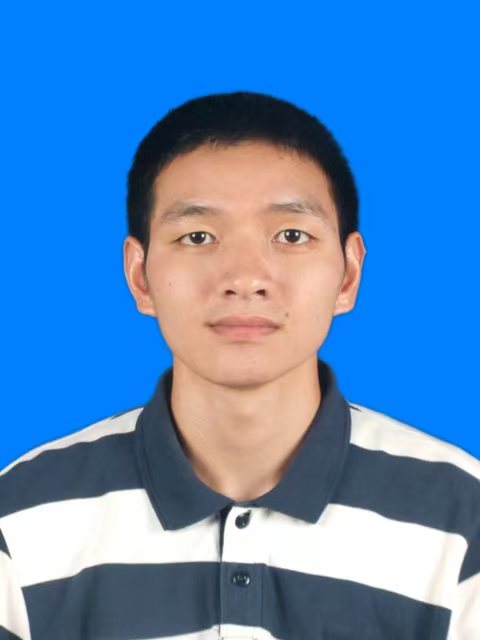}}]{Zhaoqiang Chen} 
received the B.S. and Ph.D. degrees in the School of Computer Science from Northwestern Polytechnical University, Xi'an, China. His research interests include data integration and artificial intelligence.
\end{IEEEbiography}

\begin{IEEEbiography}
[{\includegraphics[width=1in,height=1.25in,clip,keepaspectratio]{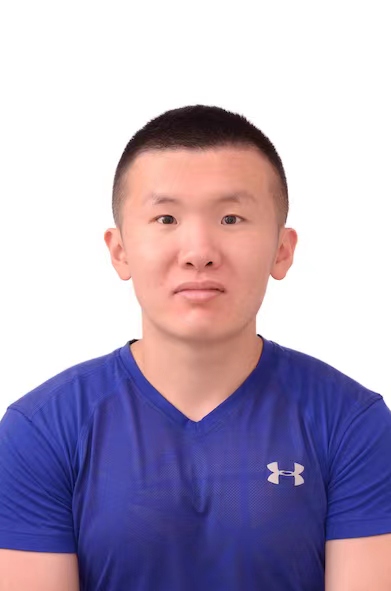}}]{Shiyu Ji}
received M.S. degree in the School of Software Engineering from Northwestern Polytechnical University, Xi'an, China. His research interests include deep learning and video understanding.
\end{IEEEbiography}

\begin{IEEEbiography}
[{\includegraphics[width=1in,height=1.25in,clip,keepaspectratio]{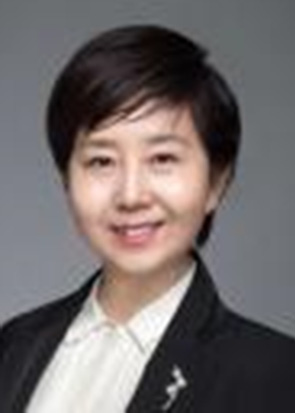}}]{Yanning Zhang} received the B.S. degree from the Dalian University of Science and Engineering in 1988 and the M.S. and Ph.D. degrees from Northwestern Polytechnical University in 1993 and 1996, respectively. She is currently a Professor with the School of Computer Science, Northwestern  Polytechnical University. She has published over 200 papers in international journals, conferences, and Chinese key journals. Her research work focuses on signal and image processing, computer vision, and pattern recognition.
\end{IEEEbiography}

\end{document}